\documentclass[]{fairmeta}

\usepackage{amsmath}
\usepackage{amssymb}
\usepackage{amsthm}
\usepackage{algorithm}
\usepackage[noend]{algpseudocode}
\usepackage{booktabs}
\usepackage{bm}
\usepackage{graphicx}
\usepackage{multirow}
\usepackage{nicematrix}
\usepackage{pifont}
\usepackage{wrapfig}
\usepackage{xcolor}
\usepackage[most]{tcolorbox}
\usepackage{hyperref}
\usepackage{url}

\titleformat*{\paragraph}{\bfseries}

\definecolor{metabg}{HTML}{F1F4F7}
\newtcolorbox{promptbox}{
    enhanced,
    breakable,
    colback=metabg,
    colframe=black!25,
    boxrule=0.4pt,
    arc=0pt,
    outer arc=0pt,
    left=8pt,
    right=8pt,
    top=8pt,
    bottom=8pt,
    before skip=6pt,
    after skip=6pt
}
\newcommand{\cmark}{\ding{51}}
\newcommand{\xmark}{\ding{55}}

\title{Edit, Don't Imitate: Minimal-Edit On-Policy Distillation for Multimodal Reasoning}

\author[1,2]{Wen Ye}
\author[1]{Ailin Deng}
\author[1,3]{Zirui Wang}
\author[1,4]{Shixing Yu}
\author[1,5]{Han Zhang}
\author[1]{Tian Yun}
\author[1]{Nikhil Mehta}
\author[1]{Tarun Kalluri}
\author[1]{Xuefei Cao}
\author[2]{Xuezhe Ma}
\author[1]{Hengduo Li}

\affiliation[1]{Meta Superintelligence Labs}
\affiliation[2]{University of Southern California}
\affiliation[3]{University of California, Berkeley}
\affiliation[4]{Cornell University}
\affiliation[5]{University of California, Los Angeles}

\abstract{Visual mathematical reasoning can fail before reasoning begins: when a vision-language model (VLM) misreads a diagram, subsequent reasoning proceeds from false premises. This perceptual bottleneck is especially pronounced in smaller models, while accurate textual descriptions can substantially narrow the gap to larger models. Because obtaining such descriptions at inference requires additional annotation or model calls, we propose minimal-edit corrective on-policy distillation. Using reference figure descriptions and verified reference solutions as training-only information, a teacher scores incorrect student rollouts under a minimal-edit objective, and only teacher-disfavored tokens contribute to the corrective loss. Correct rollouts receive task-reward reinforcement learning. This preserves on-policy supervision while avoiding full-response imitation. Across model scales, the method requires no descriptions at test time and meets or exceeds the corresponding base model supplied with a reference figure description at inference. It also improves perception-oriented performance, particularly low-level geometric perception, and transfers to external mathematics and perception benchmarks. Because the minimal-edit teacher evaluates corrections rather than generating its own solution, the method remains effective with a frozen same-size teacher, although its gains are smaller than with a larger teacher on most metrics.}

\date{\today}

\begin{document}

\maketitle

\section{Introduction}
\label{section:intro}

Visual mathematics benchmarks are commonly treated as tests of reasoning, yet many failures occur
before reasoning begins: when a vision-language model (VLM) misreads the diagram, even valid
subsequent reasoning proceeds from false premises. This perceptual bottleneck is especially
pronounced in smaller models. Giving a 2B VLM a faithful textual description of the
diagram raises its Geometry3K \citep{geometry3k} pass@1 from 0.32 to 0.46, approaching the 0.49
attained by an 8B model roughly four times its size
(Figure~\ref{fig:perception_bottleneck}). Although reasoning capacity still matters, this result
identifies perception as a major source of smaller models' disadvantage. Obtaining an accurate
description at inference, however, requires additional annotation, prompting, or model calls. We
therefore ask whether this privileged perceptual signal can be used during training to improve a
model that ultimately solves directly from the image and question.

To internalize this training-only perceptual signal, the model needs feedback on how its own
solutions go wrong. Reinforcement learning with verifiable final-answer rewards provides only
outcome-level feedback, which does not identify the source of an incorrect rollout \citep{grpo}.
Process reward models could provide finer-grained signals, but require a separately trained
evaluator with process-level supervision \citep{lightman2024verify}. On-policy
distillation instead treats the model being trained as a student and lets a teacher score the
student's own token sequence, providing dense supervision at the states the student visits
\citep{gkd}. A natural first attempt is privileged on-policy distillation: a frozen teacher sees the
reference figure description while the student receives only the image and question.

\begin{figure*}
    \centering
    \includegraphics[width=\linewidth]{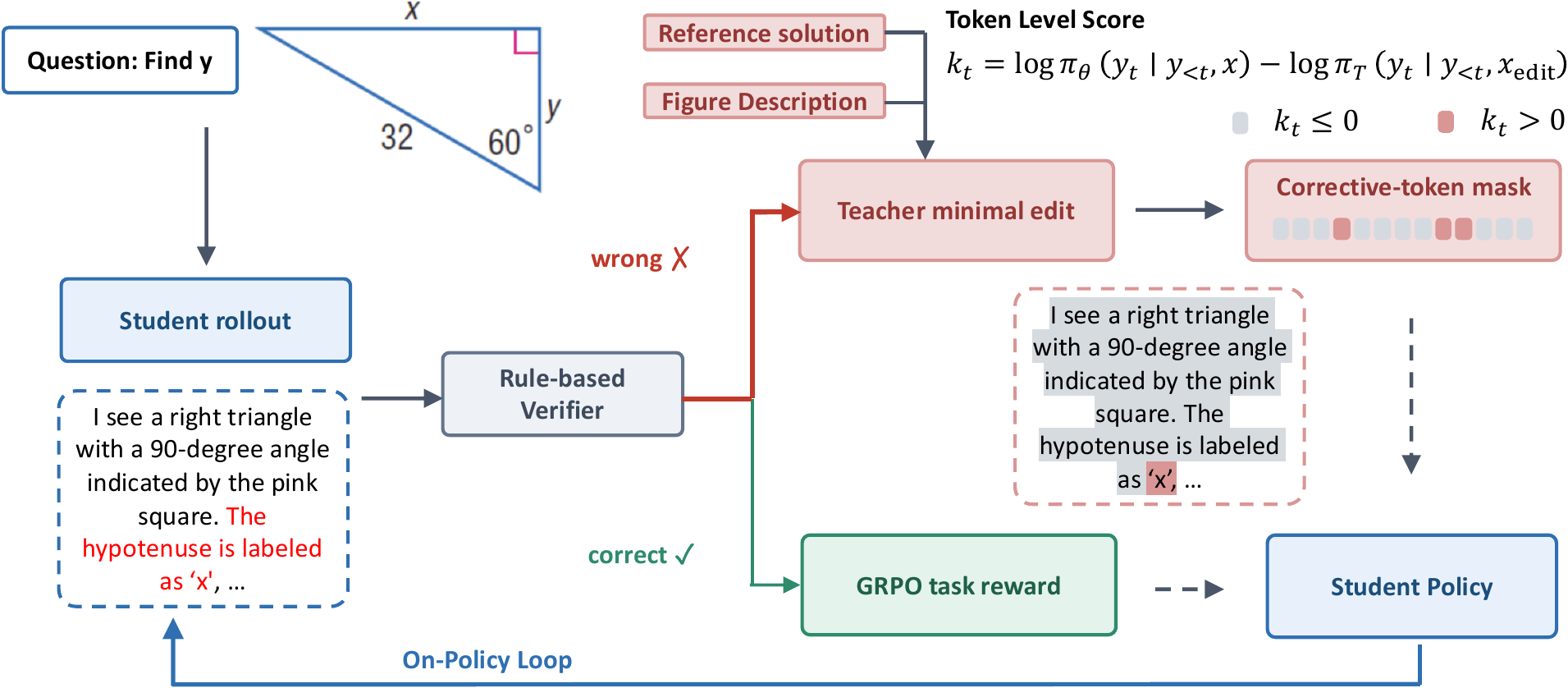}
    \caption{Overview of minimal-edit corrective on-policy distillation. A rule-based verifier routes correct rollouts to task reward; incorrect rollouts are scored by the minimal-edit teacher, and only teacher-disfavored tokens are used to update the student. Dashed arrows indicate loss calculation.}
    \label{fig:overview}
    \vspace{-0.3cm}
\end{figure*}

Standard on-policy distillation (OPD), however, applies the divergence at every token, including positions
where the teacher's preference can reflect phrasing or boilerplate rather than a substantive error
\citep{lightningopd2}. We therefore propose minimal-edit corrective on-policy distillation. For each
incorrect rollout, the minimal-edit teacher receives the reference figure description, a verified reference
solution, and the student's response. The conditioning prompt specifies a minimal-edit objective: correct all errors
while preserving the portions of the response that are already correct. The teacher does not
generate a revised solution. Instead, it scores the student's existing token sequence. Only positions where
the teacher assigns the emitted token lower probability than the student contribute to the
distillation loss and correct rollouts instead receive task-reward RL.
Figure~\ref{fig:overview} summarizes this per-rollout routing, teacher-scoring, and token selection procedure.

Scoring the student's response under a minimal-edit objective, rather than generating an independent
solution, makes the minimal-edit teacher a response-specific critic.
This less demanding role reduces the capability expected of the teacher. Experiments at 2B and 8B show that,
in the minimal-edit setting, a frozen copy of the student remains an effective teacher. Without a test-time description, minimal-edit distillation meets or exceeds
the corresponding base model evaluated with a reference figure description at inference
(Section~\ref{sec:bottleneck}), improves
performance on perception-oriented evaluations, particularly low-level geometric perception, and
generalizes to external, out-of-domain multimodal mathematics problems.

In summary, we make three contributions. First, we introduce minimal-edit corrective on-policy
distillation, which uses reference figure descriptions and verified reference solutions to select
teacher-disfavored positions in incorrect rollouts while reinforcing correct rollouts with task
reward. Second, we show that response-specific correction enables effective distillation from a
frozen same-size teacher. Third, we demonstrate that minimal-edit distillation improves reasoning
without reference figure descriptions at inference, improves perception-oriented performance, and
transfers to unseen visual-mathematics benchmarks.

\section{Related Work}
\label{sec:related}

\paragraph{On-policy and corrective distillation.} On-policy distillation is commonly optimized
with a reverse KL related to KL-regularized reinforcement learning \citep{gkd,minillm}. Multimodal
variants distill text-only teachers, steer the gradient toward visual grounding, normalize token
losses, or allocate more weight to tokens affected by fine image detail
\citep{vold,vgs,gndpo,vaopd}. Across modalities, however, token importance is
nonuniform, raw disagreement need not be learnable, and cross-teacher disagreement can reflect style
rather than task content \citep{tip,taopd,lightningopd2}. Selective objectives consequently detect
unrecoverable prefixes or use token advantages \citep{kat,aopd}. Closest to our motivation, ROSD reflects on a rollout to locate its first erroneous span, then
distills the suffix under a corrective idea \citep{rosd}; Correction2Mastery instead transfers a
generated earliest-error correction through text and a scalar reward \citep{correction2mastery}.
RLCSD contrasts correct- and incorrect-hint distributions to cancel privilege-induced style drift
\citep{rlcsd}. Our minimal-edit teacher neither generates a replacement nor defines an error suffix:
it scores the student's unchanged response under a minimal-edit instruction, and the sign of the sampled-token
log-ratio selects disfavored tokens. This differs also from methods that train an
explicit self-correction turn \citep{r3v,sherlock,octopus}, whose reliability remains limited without
external feedback \citep{huang}. More broadly, work on model self-improvement formalizes the extent
to which a model can provide useful supervision by evaluating its own generations as the
generation--verification gap \citep{mindgap}.

\paragraph{Reward routing and privileged visual supervision.} Methods combining task reward with
distillation mix the signals at response level or weight them by group difficulty
\citep{kdrl,ssopd}. SRPO instead routes correct samples to reward optimization and failed samples to
self-distillation \citep{srpo}. Our method shares this high-level partition but arrives at it from
the minimal-edit objective: only incorrect rollouts require correction, while applying the task
reward to correct rollouts avoids discarding useful samples. Our failed-rollout objective is further
distinguished by minimal-edit conditioning and token-level sign selection. A separate line of work
conditions teachers on privileged reference solutions, trajectories, or internalized contexts
\citep{lopezpaz,opsd,pidistill,opcd}. Vision-OPD uses evidence-centered crops
\citep{visionopd}; PTD-PO constructs spatial and reasoning hints for failed rollouts
\citep{ptdpo}; ViCuR uses recoverable visual cues and a cue-recovery module \citep{vicur};
and ViGOS separately supervises explicit description and reasoning spans \citep{vigos}. MaskingKD
instead hides reasoning prefixes to force reliance on the image \citep{maskingkd}. These methods
share the goal of avoiding shortcuts from privileged text: oracle-conditioned targets may be
unreachable \citep{aropd}, and when instance-specific privileged information is not recoverable
from the student's input, the student may learn an aggregate of incompatible privileged-teacher
behaviors rather than internalize the missing information \citep{manyfaces}. Our reference figure description,
by contrast, is recoverable from the image and is used only to condition a frozen
teacher; the student requires no description, cue module, or structured perception stage at
inference.

\paragraph{Perception in vision-language mathematical reasoning.} Prior work identifies perception
as a primary constraint for smaller models on diagram problems
\citep{mathverse, vlmsblind, euclid, downscaling}. Perception-aware reinforcement learning derives
training signals from image ablation \citep{papo}, consistency with visual annotations mined from
correct traces \citep{perceptionr1}, rollouts on perturbed images \citep{noisyrollout}, or a
preliminary perception stage \citep{geopqa}. Our signal instead comes from a teacher's textual
reading of the diagram; we compare against released checkpoints from this line of work on a shared backbone
(Section~\ref{sec:experiments}).

\section{Method}
\label{sec:method}

\subsection{How much do reference figure descriptions improve visual reasoning?}
\label{sec:bottleneck}

\begin{figure*}
    \centering
    \includegraphics[width=\linewidth]{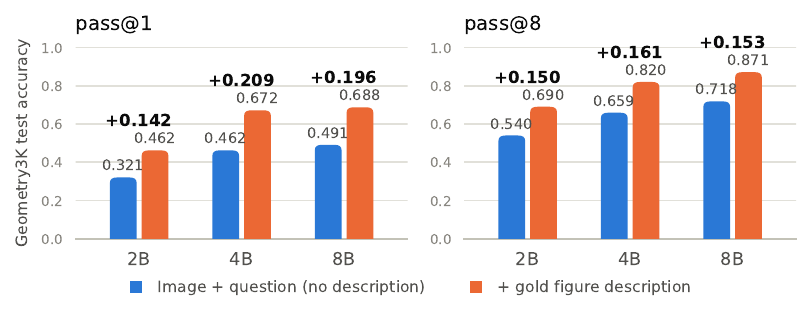}
    \caption{Geometry3K performance  with image-only input or an added reference figure description. Descriptions narrow the size gap at every scale, identifying perception as a major bottleneck.}
    \label{fig:perception_bottleneck}
    \vspace{-0.4cm}
\end{figure*}

To identify what limits small VLMs on geometry problems, we investigate how much of their performance
gap arises from diagram perception rather than downstream reasoning. We probe this distinction on
Geometry3K \citep{geometry3k} by evaluating Qwen3-VL models \citep{qwen3vl} at 2B, 4B, and 8B in
two settings: with the image and question alone, and with the same input augmented at inference by
an offline reference figure description. We generate each description with Muse Spark
\citep{meta2026musespark}, which receives the diagram, question, and ground-truth answer to check
completeness. Its prompt prohibits revealing the answer, intermediate reasoning, or derived values,
and we screen exact answer matches for leakage. Appendix~\ref{sec:fig_desc_recall} provides the full
generation and screening protocol.
Figure~\ref{fig:perception_bottleneck} shows substantial gains at every scale. With the description,
the 2B model reaches 0.462 pass@1, matching the 4B model
without it and approaching the 8B model's 0.491. The 4B model reaches 0.672, exceeding the 8B model
without a description. A faithful textual reading of the
diagram is therefore comparable to a two- to four-fold increase in parameter count in these
comparisons. Our objective is to
internalize the description's perceptual advantage so that the student approaches performance under that condition
using only the image and question.

To formalize this transfer, under on-policy distillation the student samples a rollout and the
teacher scores the same token sequence, so training remains on the student's distribution. Let
$\pi_\theta$ denote the student, $\pi_T$ the teacher, and $x$ and $x_T$ their respective inputs. For
$y\sim\pi_\theta(\cdot\mid x)$, the reverse-KL objective is

$$\mathcal{L}_{\mathrm{OPD}}(\theta) = \mathbb{E}_{y \sim \pi_\theta(\cdot \mid x)}
\sum_{t=1}^{T} \big[\log \pi_\theta(y_t \mid y_{<t}, x) - \log \pi_T(y_t \mid y_{<t},
x_T)\big],$$

where each log-ratio is the single-sample $k_1$ estimator of the conditional reverse KL at token
$t$~\citep{gkd}, and their sum estimates $\mathrm{KL}(\pi_\theta\Vert\pi_T)$ over the rollout. Unlike a scalar
rollout reward, this objective supplies a signal at each token, and its gradient coincides with
KL-regularized reinforcement learning toward the teacher. Under naive OPD, the
teacher receives the same image and question as the student and differs only in having a larger
parameter count. To use information from the reference figure description during training, we additionally give the teacher a reference figure
description $d$ while withholding it from the student, so $x_T=(d,x)$. We refer to this asymmetric
setup as privileged OPD. The student samples a rollout from the
image and question alone, and the teacher scores that same rollout with the description. The
per-token signal is the difference between the log probabilities assigned by the student and teacher
to the same token. Because the student never receives the description during training or inference,
the signal is transferred
under the input conditions the student will face at test time.
This privileged information is recoverable from the student's input: the description is a textual
reading of the image rather than an external fact. The teacher therefore supplies a capability the
student can in principle acquire from its own observation.

\subsection{Minimal-edit corrective on-policy distillation}
\label{subsec:minimaledit}
\begin{figure*}
    \centering
    \includegraphics[width=\linewidth]{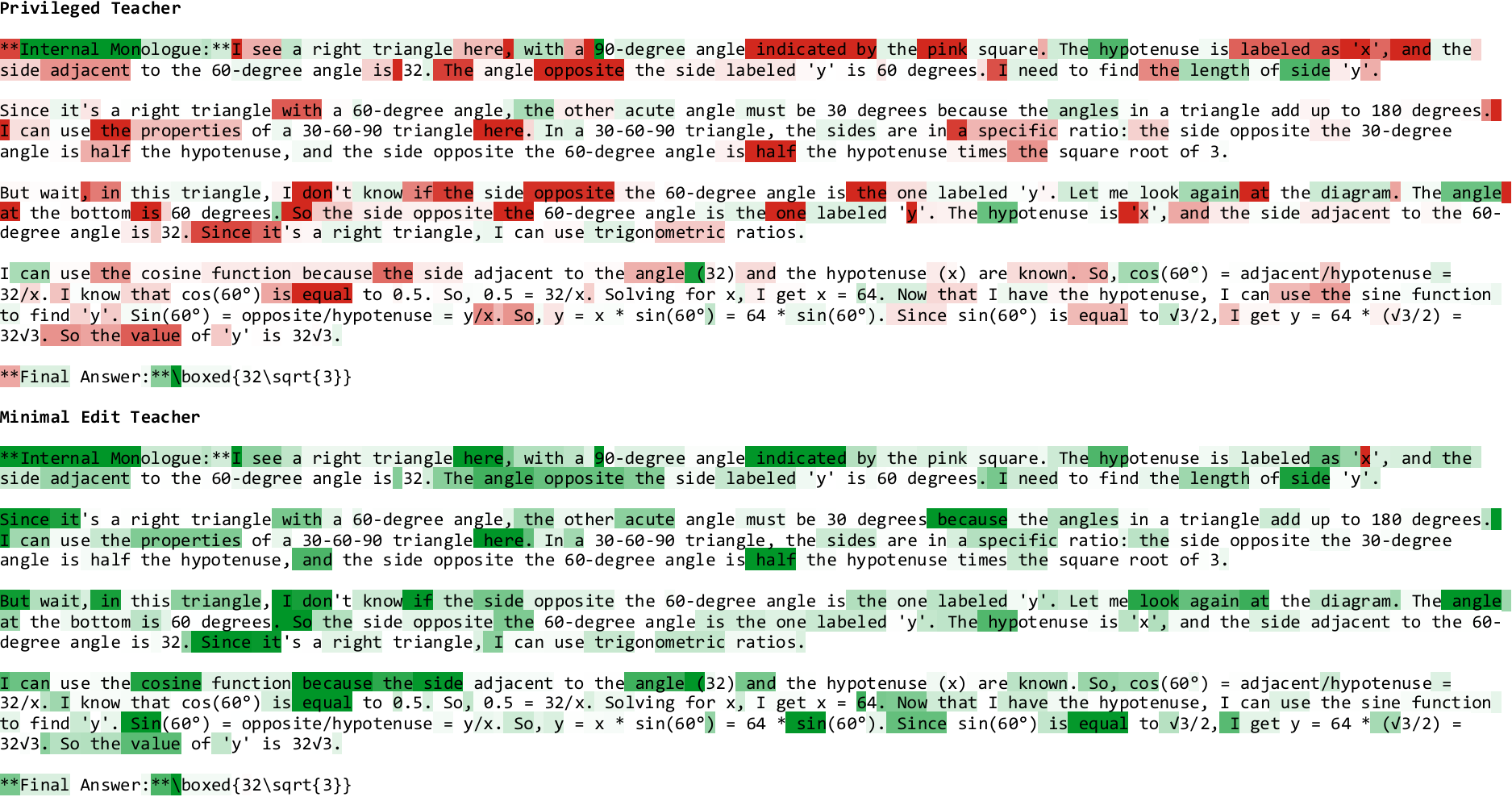}
    \caption{Token-level \(k_1\) estimates for the incorrect response to the question in Figure~\ref{fig:overview}, under privileged (top) and minimal-edit (bottom) conditioning. Red tokens (\(k_t>0\) are tokens that teacher disfavors relative to student) are retained and green tokens are discarded. Minimal-edit conditioning more precisely targets the leg–hypotenuse error, yielding a more interpretable corrective signal.}
    \label{fig:k1_visual}
    \vspace{-0.3cm}
\end{figure*}

An incorrect rollout is rarely wrong throughout: its error is often confined to a few spans, such
as a misread label, measure, or geometric relation. Applying the divergence to every token also
trains on positions where the reasoning is already correct and teacher disagreement may reflect
phrasing rather than correctness, diluting the response-specific corrective signal. For each training
problem, let $A$ denote a verified reference solution. Given an incorrect on-policy rollout $y$,
the minimal-edit teacher receives the reference figure description,
reference solution, and student rollout. It is instructed to correct all errors using the smallest
possible edits while preserving the portions that are already correct and the student's style. This
instruction defines the edit-conditioned input
$x_{\mathrm{edit}}=(d,A,y)$, but we do not generate the corrected response. Instead, the teacher
scores the student's existing sequence, yielding
$\pi_T(y_t\mid y_{<t},x_{\mathrm{edit}})$. Tokens compatible with a minimal correction should retain
high probability, whereas conflicting tokens should receive lower probability. The procedure thus
requires one teacher scoring pass and no online teacher generation. Let

$$k_t = \log \pi_\theta(y_t \mid y_{<t}, x) - \log \pi_T(y_t \mid y_{<t}, x_{\mathrm{edit}})$$

be the token-level $k_1$ estimate under edit conditioning, corresponding to the quantity summed in
the OPD objective above. If $k_t\leq0$, the minimal-edit teacher assigns at least as much
probability to the emitted token as the student does, consistent with preserving it. If $k_t>0$,
the teacher assigns less probability, which we interpret as evidence that the token conflicts with
the minimal correction. We define the binary corrective mask $m_t=\mathbf{1}[k_t>0]$ and its
selected-token set $M=\{t:m_t=1\}$; all other positions are discarded.
Figure~\ref{fig:k1_visual} illustrates the effect. The student misreads the hypotenuse of a right
triangle as a leg. The conventional privileged teacher assigns a positive $k_1$ estimate to 224 of the
515 response tokens, including many in correct passages. Under minimal-edit conditioning, the mask instead
selects a position in the phrase asserting the incorrect hypotenuse.

The minimal-edit teacher is applied only to incorrect rollouts, since correct rollouts require no
correction. Instead of discarding these useful on-policy samples, we optimize correct rollouts with
the GRPO loss. Computing advantages from only the correct rollouts would yield no relative reward
signal. We therefore first assign every rollout a binary reward using the rule-based final-answer
verifier and compute group-relative advantages over the full rollout group. We then route correct
rollouts to the GRPO loss and incorrect rollouts to the minimal-edit distillation loss. The corrective mask retains roughly 2\% of the
tokens in incorrect rollouts throughout training (Figure~\ref{fig:dynamics}). Averaging over all
response tokens would therefore shrink the corrective term and make its magnitude depend on the
number of selected positions. We instead normalize over the selected tokens:

$$\mathcal{L}(\theta) = \mathcal{L}_{\mathrm{task}}(\theta) + \frac{\lambda}{|M|} \sum_{t \in M}
\big[\log \pi_\theta(y_t \mid y_{<t}, x) - \log \pi_T(y_t \mid y_{<t}, x_{\mathrm{edit}})\big],$$

where $\mathcal{L}_{\mathrm{task}}$ is the GRPO policy-gradient objective on correct rollouts and
$\lambda$ weights the corrective term. The computational overhead relative
to standard OPD depends on the incorrect-rollout fraction and sequence length; it decreases as
training produces more correct rollouts. Appendix~\ref{sec:compute_overhead} quantifies this
tradeoff, and Appendix~\ref{sec:meopd_pseudocode} provides the complete training pseudocode.

\section{Experiments}
\label{sec:experiments}

\begin{table*}[t]
    \centering
    \caption{Main results for the Qwen3-VL-2B student. Per column, the best
    result is $\bm{bold}$ and the second best is \underline{underlined}; our minimal-edit rows are
    shaded.}
    \begin{NiceTabular}{@{}lX[c]X[c]X[c]X[c]X[c]X[c]@{}}[width=\linewidth]
    \CodeBefore
    \rectanglecolor{metabg}{13-1}{14-7}
    \Body
    \toprule
     & \multicolumn{2}{c}{Geometry3K} & MathVerse & Fig-desc & \multicolumn{2}{c}{Geoperception} \\
    \cmidrule(lr){2-3} \cmidrule(lr){6-7}
    Method & pass@1 & pass@8 & (ALL) & recall & pass@1 & pass@4 \\
    \midrule
    Qwen3-VL-2B (base)                   & $\nm{0.321}$ & $\nm{0.540}$ & $\nm{37.8}$ & $\nm{0.567}$ & $\nm{0.179}$ & $\nm{0.311}$ \\
    SFT (rejection sampling)             & $\nm{0.392}$ & $\nm{0.675}$ & $\nm{35.4}$ & $\nm{0.605}$ & $\nm{0.244}$ & $\nm{0.433}$ \\
    SFT (describe-then-solve)            & $\nm{0.475}$ & $\underline{\nm{0.791}}$ & $\nm{37.9}$ & $\nm{0.670}$ & $\nm{0.362}$ & $\nm{0.580}$ \\
    SFT (describe-figure)                & $\nm{0.214}$ & $\nm{0.598}$ & $\nm{24.3}$ & $\underline{\nm{0.770}}$ & $\nm{0.359}$ & $\nm{0.617}$ \\
    SFT (mix: rejection + describe)      & $\nm{0.406}$ & $\nm{0.688}$ & $\nm{33.5}$ & $\bm{0.784}$ & $\nm{0.289}$ & $\nm{0.473}$ \\
    GRPO (plain prompt)                  & $\nm{0.447}$ & $\nm{0.625}$ & $\nm{42.1}$ & $\nm{0.552}$ & $\nm{0.279}$ & $\nm{0.420}$ \\
    GRPO (figure-desc prompt)            & $\nm{0.173}$ & $\nm{0.361}$ & $\nm{34.9}$ & $\nm{0.415}$ & $\nm{0.278}$ & $\nm{0.474}$ \\
    GRPO (mix: plain + figure-desc)      & $\nm{0.430}$ & $\nm{0.693}$ & $\nm{43.3}$ & $\nm{0.560}$ & $\nm{0.308}$ & $\nm{0.485}$ \\
    naive OPD (8B teacher)               & $\nm{0.404}$ & $\nm{0.628}$ & $\nm{41.5}$ & $\nm{0.655}$ & $\nm{0.310}$ & $\nm{0.464}$ \\
    privileged OPD (8B teacher)          & $\nm{0.487}$ & $\nm{0.707}$ & $\nm{40.0}$ & $\nm{0.742}$ & $\nm{0.320}$ & $\nm{0.462}$ \\
    minimal-edit (8B teacher, ours)      & $\bm{0.554}$ & $\bm{0.794}$ & $\underline{\nm{45.8}}$ & $\nm{0.633}$ & $\bm{0.455}$ & $\bm{0.635}$ \\
    minimal-edit (2B self-teacher, ours) & $\underline{\nm{0.498}}$ & $\nm{0.756}$ & $\bm{46.9}$ & $\nm{0.588}$ & $\underline{\nm{0.438}}$ & $\underline{\nm{0.634}}$ \\
    \bottomrule
    \end{NiceTabular}
    \vspace{-0.3cm}
    \label{tab:main2b}
\end{table*}

\paragraph{Dataset construction}
\phantomsection\label{par:dataset_construction}
We use the Geometry3K dataset, where each problem pairs a diagram with a question and one ground-truth answer. We
use the reference figure descriptions introduced in Section~\ref{sec:bottleneck} for every training
and test diagram. In privileged and minimal-edit distillation, the student receives only
the original image and question, while the teacher additionally receives the reference figure description. Minimal-edit training additionally requires one offline reference solution per training problem.
We sample 32 solutions from the privileged teacher, conditioned on the description, figure, and
question, and retain the first whose boxed final answer passes the rule-based checker. We discard
problems for which none of the 32 candidates passes; for example, this filtering reduces the 2B
training corpus from approximately 2.1K to 1.8K problems.

\paragraph{Training setup} We evaluate three student backbones with larger teachers from the same model family:
Qwen3-VL-2B with Qwen3-VL-8B and Qwen3-VL-8B with Qwen3-VL-32B \citep{qwen3vl}, and
Qwen2.5-VL-7B with Qwen2.5-VL-32B \citep{qwen25vl}. The Qwen2.5 pair enables comparison with
published checkpoints on a shared backbone. For the Qwen3-VL-2B and Qwen3-VL-8B students, we also
run self-distillation. All larger and self-teachers are
frozen and used only for scoring. We sample $n=8$ on-policy rollouts per problem at a 2048-token response limit.
The policy-gradient term uses clipping bounds of 0.2 in both directions. We train for 30 epochs at maximum with
a learning rate of $1\times10^{-6}$ and a optimization minibatch of 96 problems per step. The
corrective weight is $\lambda=0.01$. Training is implemented in verl \citep{verl}.

\paragraph{Baselines} We compare four baseline families: (1) supervised fine-tuning on
rejection-sampled correct traces, describe-then-solve traces, the figure-description objective, or a
mixture of rejection-sampled and description traces; (2) GRPO \citep{grpo} with the plain
image-and-question prompt, the same prompt augmented with the reference figure description, or a mixture of
both; (3) naive OPD \citep{gkd}, whose teacher receives the same image and question
as the student; and (4) our privileged-OPD baseline, which additionally provides the figure
reference figure description to the teacher. No method receives the description at inference, so any benefit must be internalized in the model's
parameters. On Qwen2.5-VL-7B, we additionally compare with five published checkpoints
trained by reinforcement learning or a perception-oriented objective: Perception-R1
\citep{perceptionr1}, NoisyRollout \citep{noisyrollout}, PAPO \citep{papo}, DeepEyes
\citep{deepeyes}, and Vision-R1 \citep{visionr1}.

\paragraph{Evaluation setup.} We evaluate mathematical reasoning on Geometry3K
\citep{geometry3k} and MathVerse \citep{mathverse}, and perception with figure-description recall
and Geoperception \citep{euclid}. For mathematical reasoning, we sample 32 rollouts per Geometry3K problem at temperature 0.8 with
top-$p$ 1.0 and a 2048-token response limit, and report pass@$k$. On MathVerse, we use greedy
decoding with one response per problem, score responses with the benchmark's standard GPT-4o mini
judge \citep{openai2024gpt4omini}, and report the average over its five splits. For perception, figure-description recall measures the percentage of facts in the reference figure
description recovered by a model-generated description. We independently sample 3 descriptions
per test diagram at temperature 0.8 and report their mean recall; Appendix~\ref{sec:fig_desc_recall}
provides the complete Gemini 3.1 Pro judge protocol \citep{googledeepmind2026gemini31pro}, and Appendix~\ref{sec:detailed_perception_results} reports
the corresponding standard deviations. Geoperception tests low-level geometric relationships
organized into predicate categories of unequal size. We balance each evaluation subset by sampling
the same number of examples from every category, decode 8 rollouts per problem at temperature
0.8 with top-$p$ 1.0 and a 2048-token response limit, and average pass@$k$ over three random seeds.
Appendix~\ref{sec:detailed_perception_results} provides the predicate-level pass@1 breakdown. Tables~\ref{tab:main2b} and~\ref{tab:main8b} report results for the Qwen3-VL models, while
Table~\ref{tab:qwen25vl7b} compares methods on Qwen2.5-VL-7B.

\subsection{Main results}
\begin{table*}[t]
    \centering
    \caption{Main results for the Qwen3-VL-8B student. Per column, the best result is
    $\bm{bold}$ and the second best is \underline{underlined}; our minimal-edit rows are shaded.}
    \begin{NiceTabular}{@{}lX[c]X[c]X[c]X[c]X[c]X[c]@{}}[width=\linewidth]
    \CodeBefore
    \rectanglecolor{metabg}{13-1}{14-7}
    \Body
    \toprule
     & \multicolumn{2}{c}{Geometry3K} & MathVerse & Fig-desc & \multicolumn{2}{c}{Geoperception} \\
    \cmidrule(lr){2-3} \cmidrule(lr){6-7}
    Method & pass@1 & pass@8 & (ALL) & recall & pass@1 & pass@4 \\
    \midrule
    Qwen3-VL-8B (base)                   & $\nm{0.491}$ & $\nm{0.718}$ & $\nm{51.1}$ & $\nm{0.687}$ & $\nm{0.437}$ & $\nm{0.551}$ \\
    SFT (rejection sampling)             & $\nm{0.599}$ & $\nm{0.840}$ & $\nm{47.8}$ & $\nm{0.781}$ & $\nm{0.424}$ & $\nm{0.574}$ \\
    SFT (describe-then-solve)            & $\nm{0.691}$ & $\bm{0.911}$ & $\nm{43.5}$ & $\nm{0.798}$ & $\nm{0.500}$ & $\bm{0.686}$ \\
    SFT (describe-figure)                & $\nm{0.489}$ & $\nm{0.841}$ & $\nm{42.1}$ & $\nm{0.839}$ & $\nm{0.461}$ & $\nm{0.669}$ \\
    SFT (mix: rejection + describe)      & $\nm{0.624}$ & $\nm{0.849}$ & $\nm{50.9}$ & $\underline{\nm{0.846}}$ & $\nm{0.484}$ & $\nm{0.622}$ \\
    GRPO (plain prompt)                  & $\nm{0.664}$ & $\nm{0.823}$ & $\nm{55.1}$ & $\nm{0.676}$ & $\nm{0.530}$ & $\nm{0.648}$ \\
    GRPO (figure-desc prompt)            & $\nm{0.609}$ & $\nm{0.831}$ & $\nm{54.9}$ & $\nm{0.668}$ & $\nm{0.506}$ & $\nm{0.632}$ \\
    GRPO (mix: plain + figure-desc)      & $\nm{0.633}$ & $\nm{0.841}$ & $\nm{56.8}$ & $\nm{0.708}$ & $\nm{0.500}$ & $\nm{0.611}$ \\
    naive OPD (32B teacher)              & $\nm{0.604}$ & $\nm{0.790}$ & $\nm{57.8}$ & $\nm{0.750}$ & $\nm{0.466}$ & $\nm{0.557}$ \\
    privileged OPD (32B teacher)         & $\bm{0.708}$ & $\nm{0.837}$ & $\underline{\nm{58.2}}$ & $\bm{0.878}$ & $\nm{0.496}$ & $\nm{0.565}$ \\
    minimal-edit (32B teacher, ours)     & $\underline{\nm{0.699}}$ & $\underline{\nm{0.899}}$ & $\bm{59.0}$ & $\nm{0.727}$ & $\bm{0.553}$ & $\nm{0.673}$ \\
    minimal-edit (8B self-teacher, ours) & $\nm{0.693}$ & $\nm{0.897}$ & $\nm{53.0}$ & $\nm{0.745}$ & $\underline{\nm{0.550}}$ & $\underline{\nm{0.679}}$ \\
    \bottomrule
    \end{NiceTabular}
    \vspace{-0.4cm}
    \label{tab:main8b}
\end{table*}

\paragraph{Multimodal mathematics.} Minimal-edit distillation provides consistent gains across
Geometry3K and MathVerse. Using only the image and question at inference, both privileged and
minimal-edit distillation exceed the corresponding base model supplied with a reference figure
description, whose pass@1 is 0.462 at 2B and 0.688 at 8B. SFT remains competitive in domain but
does not transfer reliably to MathVerse, whereas GRPO transfers better but remains below
minimal-edit distillation. On the Qwen2.5-VL-7B backbone, minimal-edit distillation leads all
multimodal-math metrics while using one of the smallest training sets.

\paragraph{Perception-oriented evaluations.} Description-oriented SFT provides a strong reference
for figure-description recall because it directly optimizes that output. Minimal-edit distillation
improves both perception-oriented evaluations over the base models at 2B and 8B, whereas GRPO with
reference figure descriptions does not reliably improve perception over the base model. On Qwen2.5-VL-7B,
minimal-edit distillation also performs competitively with published external models on the
perception-oriented metrics. Perception-R1 leads figure-description recall and Geoperception pass@1, which confirms the construct validity of our perception oriented evaluations. Together, the SFT and GRPO results show that direct figure-description supervision does not consistently transfer across reasoning and perception. Most clearly for 2B figure-description-prompt GRPO, which underperforms the base model on every metric except Geoperception.

\paragraph{Minimal-edit distillation reduces dependence on teacher scale because the teacher need not generate
an independent solution.} Conditioned on the student's rollout and a verified reference, it is
instructed to make the smallest edits needed to correct the rollout. Correcting an existing solution
in this way is more verification-like than generating a solution from scratch. This interpretation
is consistent with work on the generation--verification gap, which shows that models
can derive useful supervision by evaluating their own generations once they have sufficient
verification capability \citep{mindgap}. Empirically, the self-teacher remains close to the larger
teacher at both scales. Their gap is negligible at 8B but slightly more pronounced at 2B, consistent with the
smaller model's limited verification ability. Even at 2B, however,
minimal-edit self-distillation still outperforms naive and privileged distillation with the larger teacher.

\begin{table*}[t]
    \centering
    \caption{Published Qwen2.5-VL-7B checkpoints and our method on the same backbone. Metrics
    follow Table~\ref{tab:main2b}; per column the best result is $\bm{bold}$ and the second-best
    value or tie is \underline{underlined}.}
    \begin{NiceTabular}{@{}llX[c]X[c]X[c]X[c]X[c]X[c]@{}}[width=\linewidth]
    \CodeBefore
    \rectanglecolor{metabg}{9-1}{9-8}
    \Body
    \toprule
     & & \multicolumn{2}{c}{Geometry3K} & MathVerse & Fig-desc & \multicolumn{2}{c}{Geoperception} \\
    \cmidrule(lr){3-4} \cmidrule(lr){7-8}
    Model & Train & pass@1 & pass@8 & (ALL) & recall & pass@1 & pass@4 \\
    \midrule
    Qwen2.5-VL-7B  & --            & $\nm{0.373}$ & $\nm{0.683}$ & $\nm{45.1}$ & $\nm{0.431}$ & $\nm{0.330}$ & $\underline{\nm{0.575}}$ \\
    Perception-R1-7B                 & 1.4K          & $\nm{0.504}$ & $\nm{0.690}$ & $\nm{44.9}$ & $\bm{0.541}$ & $\bm{0.420}$ & $\nm{0.570}$ \\
    NoisyRollout-7B                  & 2.1K+6.4K   & $\underline{\nm{0.522}}$ & $\underline{\nm{0.709}}$ & $\nm{47.9}$ & $\underline{\nm{0.500}}$ & $\underline{\nm{0.410}}$ & $\nm{0.556}$ \\
    PAPO-7B                          & 38.8K         & $\nm{0.407}$ & $\nm{0.640}$ & $\underline{\nm{48.9}}$ & $\nm{0.470}$ & $\nm{0.351}$ & $\nm{0.540}$ \\
    DeepEyes-7B                      & 47K           & $\nm{0.393}$ & $\nm{0.693}$ & $\nm{46.1}$ & $\nm{0.447}$ & $\nm{0.363}$ & $\bm{0.588}$ \\
    Vision-R1-7B                     & 200K+10K    & $\nm{0.436}$ & $\nm{0.637}$ & $\nm{47.7}$ & $\nm{0.486}$ & $\nm{0.392}$ & $\nm{0.552}$ \\
    minimal-edit & 1.8K          & $\bm{0.543}$ & $\bm{0.718}$ & $\bm{49.3}$ & $\nm{0.477}$ & $\underline{\nm{0.410}}$ & $\nm{0.568}$ \\
    \bottomrule
    \end{NiceTabular}
    \vspace{-0.3cm}
    \label{tab:qwen25vl7b}
\end{table*}

\begin{figure*}
    \centering
    \includegraphics[width=\linewidth]{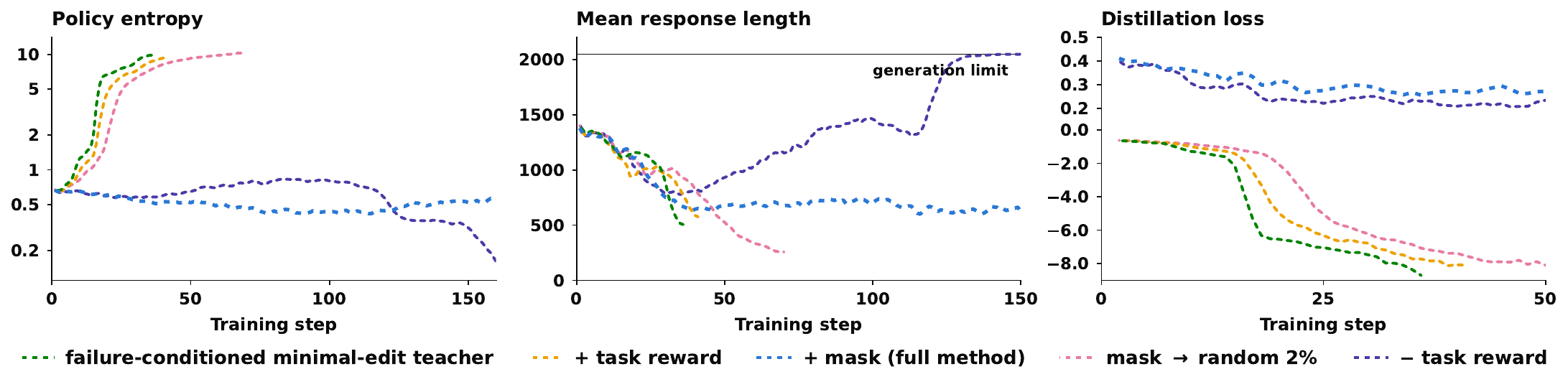}
    \caption{Training dynamics for Qwen3-VL-2B component ablations: policy entropy (left), mean
    response length (center), and distillation loss (right). Only the full method remains stable. }
    \label{fig:component_ablation}
    \vspace{-0.3cm}
\end{figure*}

\subsection{Ablation studies}
\label{subsec:ablations}

Table~\ref{tab:coefsweep} tests the sensitivity of minimal-edit distillation to the corrective
weight $\lambda$ and to the verified reference solution. Using the Qwen3-VL-2B student and
Qwen3-VL-8B teacher, we vary $\lambda$ from $0.001$ to $0.1$ while holding the rest of the training
configuration fixed; the final row removes the reference solution from the selected
$\lambda=0.01$ configuration. Every swept weight improves on the untrained base across all measurements. The corrective weight has
a nonmonotonic effect: an intermediate weight performs consistently best, whereas either weakening
or strengthening the distillation term reduces accuracy. This suggests that minimal-edit
supervision is most effective when it complements rather than overwhelms the task-reward signal.
Removing the verified reference solution also degrades all six measures. The reference solution
therefore serves as an additional anchor for identifying how an incorrect rollout should be
corrected.  \begin{wraptable}{r}{0.5\textwidth}
    \vspace{-0.3cm}
    \centering
    \footnotesize
    \setlength{\tabcolsep}{5pt}
        \caption{Component configurations used in the ablation study.}
    \begin{NiceTabular}{@{}lccc@{}}
    \CodeBefore
    \rectanglecolor{metabg}{6-1}{6-4}
    \Body
    \toprule
    & ME Teacher & Reward & Mask \\
    \midrule
    privileged OPD            & \xmark & \xmark & \xmark \\
    + ME teacher           & \cmark & \xmark & \xmark \\
    \quad + task reward       & \cmark & \cmark & \xmark \\
    \quad + random 2\% mask   & \cmark & \cmark & $\mathrm{R}$ \\
    full method               & \cmark & \cmark & \cmark \\
    \quad $-$ task reward     & \cmark & \xmark & \cmark \\
    \bottomrule
    \end{NiceTabular}

    \label{tab:ablconfig}
    \vspace{-0.8cm}
\end{wraptable}This ablation does not separately isolate the figure description because reference
solutions are themselves generated with it to maximize verified-solution yield. Removing it only
during teacher scoring leaves this dependence intact, while regenerating references without
descriptions also changes solution yield and quality.

Table~\ref{tab:ablconfig} defines the component combinations. ME Teacher denotes minimal-edit
conditioning for scoring incorrect rollouts, Reward applies task reward to correct rollouts, and
Mask refers to the corrective mask that retains tokens with $k_t>0$; $\mathrm{R}$ instead denotes a uniformly random token position mask that matches
the corrective mask's empirical keep rate of roughly 2\% (Figure~\ref{fig:dynamics}). Components are
added cumulatively through task reward, after which the random-mask control and the full method
differ only in how masked tokens are selected. The final row removes task reward from the full
configuration. Runs without task reward contain no combined loss and therefore require no corrective
weight; all other ablation runs use $\lambda=0.01$.
Figure~\ref{fig:component_ablation} shows the corresponding training dynamics. The incomplete
configurations become unstable through two distinct failure modes.
Under minimal-edit conditioning, the teacher sees the student's rollout and is explicitly
instructed to preserve its correct portions (Appendix~\ref{sec:minimal_edit_prompt}). Higher teacher
probability on these already supplied tokens therefore largely reflects copying from the rollout
rather than corrective information that the student should learn.
Because preserved tokens constitute most of the response, their negative
token-level $k_1$ estimates dominate the unmasked objective (Figure~\ref{fig:k1_visual}). The
distillation term consequently becomes strongly negative and policy entropy explodes. Replacing
the corrective mask with the random 2\% mask produces the same failure mode. Matching the mask's
sparsity is therefore insufficient and validates that the tokens selected
by the corrective mask carry meaningful corrective information.

Without task reward on correct rollouts, the opposite failure occurs: entropy collapses and
responses grow to the generation limit. This suggests that task reward does more than retain
otherwise unused correct samples: by reinforcing successful on-policy behavior, it anchors the
policy to solution patterns already known to succeed. This interpretation is consistent with the
long-horizon stability obtained by routing correct samples to reward-aligned RL in SRPO
\citep{srpo}, and with evidence that verified-correct reference rollouts can prevent policy drift
\citep{armor}. Both failure modes drive a marked decline in validation accuracy and result in
optimization collapse. By contrast, the full method maintains stable entropy and distillation loss
throughout training.
\begin{table*}
        \centering
            \caption{Corrective-weight sweep and reference-solution ablation for the Qwen3-VL-2B student
    with a Qwen3-VL-8B teacher. Performance peaks at an intermediate weight and declines without
    the reference solution. Per column, the best result is $\bm{bold}$ and the second best is
    \underline{underlined}; the configuration used in our main experiments is shaded.}
    \begin{NiceTabular}{@{}lX[c]X[c]X[c]X[c]X[c]X[c]@{}}[width=\linewidth]
    \CodeBefore
    \rectanglecolor{metabg}{5-1}{5-7}
    \Body
    \toprule
     & \multicolumn{2}{c}{Geometry3K} & MathVerse & Fig-desc & \multicolumn{2}{c}{Geoperception} \\
    \cmidrule(lr){2-3} \cmidrule(lr){6-7}
    $\lambda$ & pass@1 & pass@8 & (ALL) & recall & pass@1 & pass@4 \\
    \midrule
    $0.001$ & $\nm{0.442}$ & $\nm{0.736}$ & $\nm{42.5}$ & $\nm{0.584}$ & $\nm{0.373}$ & $\nm{0.550}$ \\
    $0.005$ & $\nm{0.432}$ & $\nm{0.721}$ & $\nm{39.0}$ & $\nm{0.589}$ & $\nm{0.374}$ & $\nm{0.569}$ \\
    $0.01$  & $\bm{0.554}$ & $\bm{0.794}$ & $\bm{45.8}$ & $\bm{0.633}$ & $\bm{0.455}$ & $\bm{0.635}$ \\
    $0.05$  & $\nm{0.407}$ & $\nm{0.697}$ & $\nm{42.3}$ & $\nm{0.609}$ & $\nm{0.396}$ & $\nm{0.594}$ \\
    $0.1$   & $\nm{0.366}$ & $\nm{0.655}$ & $\nm{41.1}$ & ${\nm{0.610}}$ & $\nm{0.357}$ & $\nm{0.562}$ \\
    \midrule
    $0.01$ (no ref. solution) & $\underline{\nm{0.471}}$ & $\underline{\nm{0.752}}$ & $\underline{\nm{43.8}}$ & $\underline{\nm{0.618}}$ & $\underline{\nm{0.433}}$ & $\underline{\nm{0.625}}$ \\
    \bottomrule
    \end{NiceTabular}
    \vspace{-0.5cm}
    \label{tab:coefsweep}
\end{table*}

\section{Conclusion and Future Perspectives}
\label{sec:conclusion}

In this work, we identify diagram perception as a major source of smaller VLMs' disadvantage in
visual mathematical reasoning and introduce minimal-edit corrective on-policy distillation to
use privileged perceptual information for response-specific corrective supervision without
requiring it at inference. For each incorrect
rollout, a privileged minimal-edit teacher scores the student's existing tokens, and only positions
the teacher disfavors are used to update the policy; correct rollouts instead receive task reward.
Across model scales, this response-specific supervision improves reasoning and performance on
perception-oriented evaluations, particularly low-level geometric perception, meets or exceeds the
corresponding base model evaluated with a reference figure description at inference, and remains effective
with a frozen same-size teacher. By comparison, supervised fine-tuning,
description-conditioned reinforcement learning, and full-token privileged distillation do not
transfer the benefit of figure descriptions as consistently.

More broadly, our findings suggest that recoverable privileged information can be used as a
training-time diagnostic rather than an additional input or generation target. Disagreement with a
privileged critic can expose where information available in a complex visual input fails to enter
the model's reasoning. One limitation is that the sign test used by the reverse-KL mask is not
guaranteed to identify every token occurrence that reflects an underlying error. In
Figure~\ref{fig:k1_visual}, for example, some occurrences of the student's leg--hypotenuse
misreading do not receive positive $k_1$ estimates and are therefore not selected by the mask.
Nevertheless, this partial localization provides an effective corrective signal empirically in our experiments.
A future direction of this work may extend beyond geometry to charts, scientific figures,
tables, and other domains in which an additional textual reading makes visual evidence easier to
verify. This will require reliable ways to construct privileged views that are
faithful to the original input, aligned with the task, and informative about the specific errors a
model can learn to correct.

\bibliographystyle{assets/plainnat}
\bibliography{paper}

\clearpage
\newpage
\beginappendix

\section{Minimal-Edit Corrective OPD Pseudocode}
\label{sec:meopd_pseudocode}

Algorithm~\ref{alg:meopd} gives the complete training procedure. Each training example contains the
student input $x_i$ (image and question), a training-only figure description $d_i$, a verified
reference solution $A_i$, and the ground-truth answer $g_i$. The verifier $V$ maps a rollout and
$g_i$ to a binary correctness reward. We first sample all $K$ rollouts from a frozen snapshot of
the current student and compute their task advantages from the $K$ verifier rewards. A correct
rollout enters only the task-reward branch. For an incorrect rollout, the teacher receives the
complete rollout inside the minimal-edit prompt and scores a second, teacher-forced occurrence of
the same rollout. Thus, \textsc{TeacherForce} is a scoring pass, not a decoding call.

The corrective mask is computed from detached student and teacher scores and is therefore fixed
during the subsequent policy update. We normalize the corrective term by the number $Z$ of selected
tokens across the minibatch; if $Z=0$, its contribution is zero. In the experiments,
\textsc{TaskAdvantages} and $\widehat{\mathcal{L}}_{\mathrm{PG}}$ are the group-relative advantages and
clipped policy-gradient objective described in Section~\ref{sec:experiments}. The task objective is
evaluated only on correct rollouts, although its group statistics are computed from all $K$
verifier rewards. Consequently, every sampled rollout contributes through exactly one branch.

\begin{algorithm}
\caption{Minimal-edit corrective on-policy distillation (ME-OPD)}
\label{alg:meopd}
\begin{algorithmic}[1]
\Require Dataset $\mathcal{D}=\{(x_i,d_i,A_i,g_i)\}$; student $\pi_\theta$; frozen teacher
$\pi_T$; verifier $V$; rollouts per example $K$; corrective weight $\lambda$
\For{each optimization step}
    \State Sample a minibatch $\mathcal{B}\subset\mathcal{D}$ and set $\theta_{\mathrm{old}}\gets\theta$
    \State Initialize correct-rollout buffer $\mathcal{C}\gets\varnothing$ and incorrect-rollout
    buffer $\mathcal{I}\gets\varnothing$
    \For{each $(x_i,d_i,A_i,g_i)\in\mathcal{B}$}
        \For{$j=1,\ldots,K$}
            \State $y_{ij}\sim\pi_{\theta_{\mathrm{old}}}(\cdot\mid x_i)$
            \State $r_{ij}\gets V(y_{ij},g_i)\in\{0,1\}$
        \EndFor
        \State $(\widehat a_{i1},\ldots,\widehat a_{iK})\gets
        \Call{TaskAdvantages}{r_{i1},\ldots,r_{iK}}$
        \For{$j=1,\ldots,K$}
            \If{$r_{ij}=1$}
                \State Append $(x_i,y_{ij},\widehat a_{ij})$ to $\mathcal{C}$
            \Else
                \State $x^{ij}_{\mathrm{edit}}\gets
                \Call{EditPrompt}{x_i,d_i,A_i,y_{ij}}$
                \State $\ell^{S}_{ij}\gets
                \Call{LogProbs}{\pi_{\theta_{\mathrm{old}}},x_i,y_{ij}}$
                \Comment{student scores from the rollout-policy}
                \State $\ell^{T}_{ij}\gets
                \Call{TeacherForce}{\pi_T,x^{ij}_{\mathrm{edit}},y_{ij}}$
                \Comment{teacher scores the same rollout}
                \State $m_{ij}\gets\operatorname{stopgrad}\!\left(
                \mathbf{1}[\ell^{S}_{ij}-\ell^{T}_{ij}>0]\right)$
                \Comment{compute the corrective mask}
                \State Append $(x_i,y_{ij},\ell^{T}_{ij},m_{ij})$ to $\mathcal{I}$
            \EndIf
        \EndFor
    \EndFor
    \If{$|\mathcal{C}|>0$}
        \State $\widehat{\mathcal{L}}_{\mathrm{task}}\gets
        \widehat{\mathcal{L}}_{\mathrm{PG}}(\theta;\theta_{\mathrm{old}},\mathcal{C})$
    \Else
        \State $\widehat{\mathcal{L}}_{\mathrm{task}}\gets0$
        \Comment{no correct rollouts}
    \EndIf
    \State $Z\gets\displaystyle\sum_{(x,y,\ell^T,m)\in\mathcal{I}}\sum_t m_t$
    \Comment{number of selected corrective tokens}
    \If{$Z>0$}
        \State $\widehat{\mathcal{L}}_{\mathrm{corr}}\gets\displaystyle\frac{1}{Z}
        \sum_{(x,y,\ell^T,m)\in\mathcal{I}}\sum_t m_t
        \left[\log\pi_\theta(y_t\mid y_{<t},x)-\ell^T_t\right]$
    \Else
        \State $\widehat{\mathcal{L}}_{\mathrm{corr}}\gets0$
        \Comment{avoid division by zero when no token is selected}
    \EndIf
    \State $\widehat{\mathcal{L}}\gets
    \widehat{\mathcal{L}}_{\mathrm{task}}+\lambda
    \widehat{\mathcal{L}}_{\mathrm{corr}}$
    \State Update $\theta$ by descending $\nabla_\theta\widehat{\mathcal{L}}$
\EndFor
\end{algorithmic}
\end{algorithm}

At inference, the trained student receives only $x$; the reference figure description, reference solution,
teacher, verifier, and edit prompt are all training-only components.

\section{Computational Cost Analysis}
\label{sec:compute_overhead}

We compare the per-step floating-point operation (FLOP) cost of minimal-edit corrective on-policy
distillation with naive OPD, holding student behavior fixed at the sequence lengths and
incorrect-rollout fractions observed in the minimal-edit runs. Let $N$ be the number of student parameters and let the teacher contain $rN$ parameters. Let $P_s$
and $R$ denote the mean student-prompt and response lengths, respectively, and define the complete
student sequence length as $D=P_s+R$. Each problem produces $K$ student rollouts, of which a
fraction $w$ are incorrect. Finally, $P_t$ denotes the fixed portion of the minimal-edit teacher
prompt in Section~\ref{sec:minimal_edit_prompt}, including the edit instruction, figure description,
reference solution, and image tokens, measured with an empty student response.

\citet{kaplan2020scaling} give a leading-order forward cost of $2N$ FLOPs per token, apart from a
context-length-dependent attention term; over $D$ tokens, this yields $2ND$.
\citet{hoffmann2022training} use $6ND$ as the corresponding forward--backward training cost. Our
runs use activation recomputation (gradient checkpointing), which re-evaluates forward activations
during backpropagation and raises the update cost to $8ND$.

Both methods share the same student computation: rollout generation over $R$ response tokens,
a separate post-rollout forward pass that evaluates behavior-policy log probabilities over $D$
tokens, and a policy update over $D$ tokens. Their respective costs are
\begin{equation}
    C_{\mathrm{gen}} \simeq 2KNR, \qquad
    C_{\mathrm{logprob}} \simeq 2KND, \qquad
    C_{\mathrm{train}} \simeq 8KND,
\end{equation}
and hence
\begin{equation}
    C_{\mathrm{student}} \simeq KN(2R+10D).
\end{equation}

In naive OPD, the teacher scores every student rollout over the complete sequence of length $D$.
The teacher cost is $2rKND$, giving
\begin{equation}
    \begin{aligned}
    C_{\mathrm{naive}}
    &\simeq KN\left[2R+(10+2r)D\right] \\
    &\simeq KND\left[\frac{2R}{D}+10+2r\right].
    \end{aligned}
    \label{eq:compute_naive}
\end{equation}

The minimal-edit teacher scores only the fraction $w$ of rollouts that are incorrect. For each
teacher-scored rollout, the on-policy student response, with mean length $R$, occurs once in the
conditioning prompt and once as the suffix whose log probabilities are evaluated, so the teacher
processes $P_t+2R$ tokens. One copy is embedded in the teacher prompt, allowing the teacher to
condition on the complete student response; the other is the teacher-forced suffix whose log
probabilities are evaluated.
The resulting total is
\begin{equation}
    \begin{aligned}
    C_{\mathrm{edit}}
    &\simeq KN\left[2R+10D+2wr(P_t+2R)\right] \\
    &\simeq KND\left[\frac{2R}{D}+10+
    \frac{2wr(P_t+2R)}{D}\right].
    \end{aligned}
    \label{eq:compute_edit}
\end{equation}
The corrective mask and correctness gate require no additional model evaluations.

Dividing Eq.~\eqref{eq:compute_edit} by Eq.~\eqref{eq:compute_naive} yields
\begin{equation}
    \frac{C_{\mathrm{edit}}}{C_{\mathrm{naive}}}
    = 1 +
    \frac{2r\left[w(P_t+2R)-D\right]}
    {2R+(10+2r)D}.
    \label{eq:compute_ratio}
\end{equation}
Because the denominator is positive, minimal-edit distillation is less expensive than naive OPD
whenever
\begin{equation}
    w < w^* = \frac{D}{P_t+2R}.
    \label{eq:compute_break_even}
\end{equation}
The right-hand side is the ratio of the sequence length processed by the naive teacher for each
rollout, $D$, to the sequence length processed by the minimal-edit teacher for each scored rollout,
$P_t+2R$. Thus, at $w=w^*$, the expected teacher-processed tokens per rollout are equal:
$w^*(P_t+2R)=D$.

Table~\ref{tab:compute_inputs} reports the quantities used in the comparison. Response length,
prompt length, and incorrect-rollout fraction are collected from the Weights \& Biases (W\&B) run
logs and averaged over all training steps. Using the minimal-edit statistics for both methods
isolates the training recipe from differences in response length and truncation behavior.

\begin{table*}[t]
    \centering
    \caption{Inputs to the computational-cost estimate. Sequence lengths are measured in tokens.}
    \label{tab:compute_inputs}
    \small
    \begin{NiceTabular}{@{}lX[c]X[c]@{}}[width=0.78\textwidth]
        \toprule
        Quantity & 8B student / 32B teacher & 2B student / 8B teacher \\
        \midrule
        Mean response length $R$ & 736.62 & 783.46 \\
        Mean student-prompt length $P_s$ & 209.32 & 209.35 \\
        Student sequence $D=P_s+R$ & 945.94 & 992.81 \\
        Incorrect-rollout fraction $w$ & 0.275 & 0.456 \\
        Parameter ratio $r$ & 4.00 & 4.00 \\
        Fixed edit-prompt length $P_t$ & 1369.2 & 1316.6 \\
        \bottomrule
    \end{NiceTabular}
\end{table*}

\begin{table*}[t]
    \centering
    \caption{Estimated per-step compute of minimal-edit distillation relative to naive OPD.}
    \label{tab:compute_results}
    \scriptsize
    \begin{NiceTabular}{@{}lX[c]X[c]X[c]X[c]X[c]X[c]@{}}[width=\textwidth]
        \toprule
        Configuration & $D$ & $P_t+2R$ & $w^*$ & $C_{\mathrm{naive}}$ & $C_{\mathrm{edit}}$ & $C_{\mathrm{edit}}/C_{\mathrm{naive}}$ \\
        \midrule
        8B student / 32B teacher & 945.94 & 2842.44 & 0.333 & $19.56\,KND$ & $18.17\,KND$ & \textbf{0.929} \\
        2B student / 8B teacher & 992.81 & 2883.52 & 0.344 & $19.58\,KND$ & $22.17\,KND$ & \textbf{1.133} \\
        \bottomrule
    \end{NiceTabular}
\end{table*}

Substituting the measured values into
Eqs.~\eqref{eq:compute_naive}--\eqref{eq:compute_break_even} gives
Table~\ref{tab:compute_results}. We report each cost as a multiple of $KND$; this common factor
cancels in the ratio. For the 8B student, $w=0.275<w^*=0.333$, so the minimal-edit teacher processes
$w(P_t+2R)=781.67$ tokens per rollout in expectation, fewer than the $D=945.94$ tokens processed by
the naive teacher. Minimal-edit distillation therefore requires $7.1\%$ fewer FLOPs. For the 2B
student, $w=0.456>w^*=0.344$ and $w(P_t+2R)=1314.89>D=992.81$, yielding $13.3\%$ more FLOPs.

Figure~\ref{fig:dynamics} shows that the correct-rollout fraction increases over the 2B run, so the
complementary incorrect-rollout fraction $w$ decreases. The minimal-edit teacher is therefore
invoked less often as training progresses, reducing the cost of minimal-edit OPD relative to naive
OPD, whose teacher scores every rollout.

\section{Training Dynamics of the Minimal-Edit Run}
\label{subsec:full_dynamics}

\begin{figure*}
    \centering
    \includegraphics[width=\linewidth]{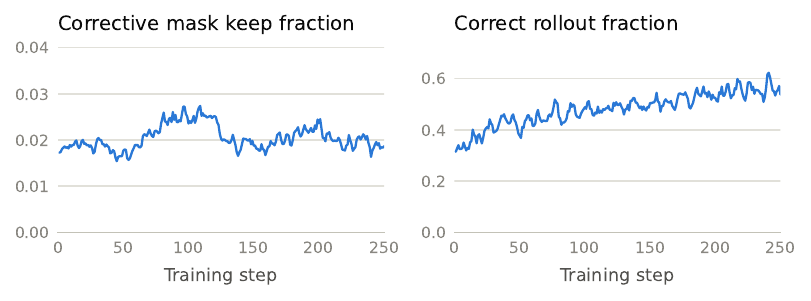}
    \caption{Training dynamics of the full minimal-edit run. Corrective-mask keep rate on incorrect rollouts (left) and correct-rollout fraction (right). The mask stays near 2\% as accuracy rises, shifting training toward task reward without diffusing corrective supervision.}
    \label{fig:dynamics}
\end{figure*}

Figure~\ref{fig:dynamics} reports the two quantities that determine how supervision is allocated
during the full Qwen3-VL-2B minimal-edit run. Among incorrect rollouts, the corrective mask retains
only about $2\%$ of response tokens. The keep fraction fluctuates within a narrow band over 250
steps and shows no sustained upward trend, even as the on-policy distribution changes. The teacher
therefore continues to identify a small set of candidate corrections.

At the same time, the correct-rollout fraction rises from roughly one third early in training to
more than one half near the end. Because correct and incorrect rollouts are handled by mutually
exclusive branches, this increase changes the mixture of supervision: more sampled responses
receive task-reward reinforcement, while the remaining failures continue to receive sparse
token-level correction. The joint behavior is consistent with the intended division of labor in
the objective. The model produces successful rollouts more often without losing the sparse
character of the corrective signal on unsuccessful ones.

\section{Baseline Training Details}
\label{sec:baseline_training_details}
\paragraph{Supervised fine-tuning.} We train all supervised baselines with LlamaFactory
\citep{llamafactory} using full-parameter fine-tuning: the vision encoder, multimodal projector, and
language model are all unfrozen. Training uses AdamW \citep{adamw} in bfloat16 with DeepSpeed
ZeRO-3 \citep{zero} for up to three epochs, a learning rate of $1\times10^{-5}$ with cosine decay, a
warmup ratio of 0.1, and an effective batch size of 12 sequences. The maximum combined
prompt-and-target length is 4096 tokens. Input images are capped at 262144 pixels, corresponding to
at most 334 visual tokens per figure.

\paragraph{Reinforcement learning.} We train the reinforcement-learning baselines with GRPO
\citep{grpo} in verl \citep{verl}, using an FSDP2 actor \citep{fsdp} and vLLM \citep{vllm} for
rollout generation. Each step samples eight rollouts per prompt at temperature 1.0 from a batch of
96 prompts. We use a minibatch of 96 prompts and one optimization epoch per rollout batch, yielding
a single strictly on-policy update. Prompts and responses are limited to 1024 and 2048 tokens,
respectively. The actor is optimized with AdamW \citep{adamw} using a learning rate of
$1\times10^{-6}$, a constant schedule without warmup, weight decay of 0.01, and gradient clipping at
1.0. We use a symmetric clipping range of 0.2, a dual-clip bound of 3.0, no entropy bonus, and
token-mean loss aggregation. We apply neither an explicit KL loss nor a KL penalty in the reward and
therefore retain no reference model. The reward is the rule-based Geometry3K correctness signal,
and training runs for 30 epochs. For both supervised fine-tuning and GRPO, we select the checkpoint
with the highest validation accuracy.

\section{Reference Figure Description Collection and Use}
\label{sec:fig_desc_recall}

\subsection{Figure-description collection}

\begin{figure*}
    \centering
    \includegraphics[width=\linewidth]{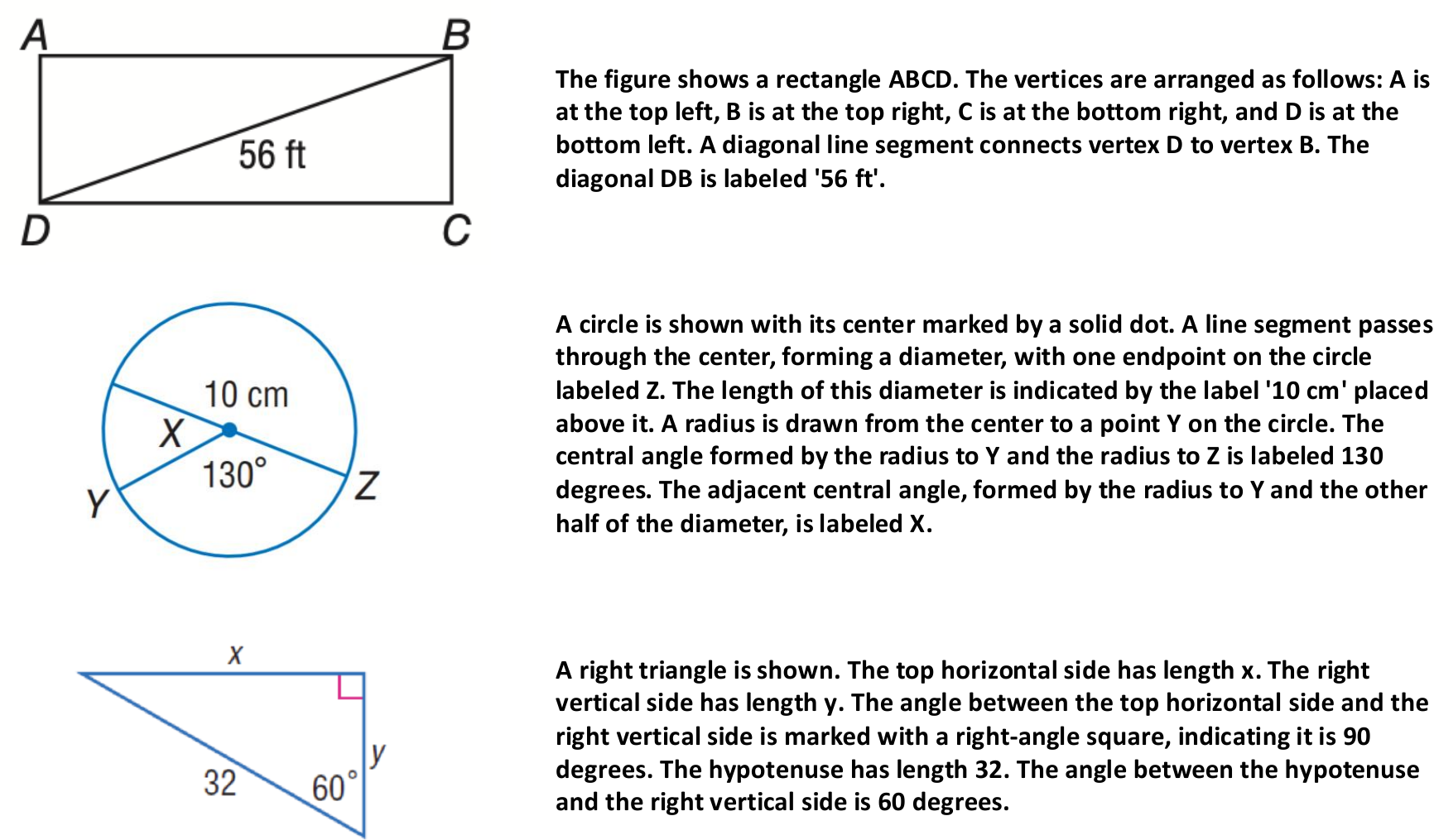}
    \caption{Qualitative examples of reference figure descriptions used for the
    figure-description recall evaluation. Each row pairs a Geometry3K diagram with its reference figure
    description, whose objects, labels, measures, and geometric relations are decomposed into
    atomic facts by the Gemini 3.1 Pro judge.}
    \label{fig:example_descriptions}
\end{figure*}

We use Muse Spark to collect the reference figure descriptions with the prompt below. For each
Geometry3K example, the model receives the diagram, the original question, and the verified final
answer. The generation prompt explicitly prohibits derived values, intermediate reasoning, or the
answer itself. These collected descriptions are the privileged visual context
provided to both the conventional privileged teacher and the minimal-edit teacher during training;
they also serve as the references for the figure-description recall evaluation below.
Figure~\ref{fig:example_descriptions} shows qualitative examples collected with this procedure.

\begin{promptbox}
\footnotesize\ttfamily\raggedright
SYSTEM INSTRUCTION (SYSTEM\_INSTRUCTION), passed via \texttt{--system}
\par
\medskip
You convert a geometry diagram into a precise textual description so that a text-only language
model, which cannot see the image, can solve the problem.
\par
\medskip
The text-only solver will receive exactly: YOUR figure description, followed by the original
question (verbatim). Your job is to make sure that combination contains every fact needed to solve
it.
\par
\medskip
You are given the diagram image, the question, and the correct final answer. Use the answer ONLY to
check completeness:
\par
-- Actually solve the problem yourself from the figure and the question.
\par
-- Confirm the answer is derivable from [your figure description + the question] ALONE. If you
cannot derive it, your description is missing a visual fact---find it in the figure and add it.
\par
-- NEVER reveal the answer, the solution, intermediate steps, or any computed/derived value in your
description. The solver must solve it itself.
\par
\medskip
Describe ONLY what is visually shown in the figure, precisely:
\par
-- Every labeled point and how the points are arranged (which are collinear, which form the vertices
of which shape).
\par
-- Every line, ray, segment, arc, circle, polygon, and intersection (intersection points, endpoints,
which segments make up a path).
\par
-- EVERY numeric measure or algebraic expression printed in the figure: side lengths, segment
lengths, angle measures, arc measures, coordinates, and labels like ``3x-1''; state which element
each value belongs to.
\par
-- Every geometric mark: right-angle squares (state which angle is 90 degrees), tick marks
indicating congruent segments, arc marks indicating congruent angles, parallel arrows, tangent
points, and center marks.
\par
-- Relationships shown in the figure: tangency, parallelism, perpendicularity, midpoints, and
bisectors---but only when the figure itself marks them.
\par
\medskip
Note on the question text: the question sometimes states constraints or definitions (e.g.,
``ZQ = 3a - 11'', ``BG is a median of triangle ABC'', ``the radius is 4''). That information stays
in the final prompt because the question is preserved verbatim, so you do not need to repeat it---but
DO account for it when checking solvability, and make sure any fact that lives ONLY in the figure
ends up in your description.
\par
\medskip
Strict rules:
\par
-- Describe only what is drawn or labeled. Do not include any value not explicitly shown in the
figure that you computed during your own solution.
\par
-- Do NOT state the answer in the description.
\par
-- If a value or label is unreadable, write ``unlabeled'' rather than guessing.
\par
-- Use plain text and standard math notation (e.g., ``angle ABC = 40 degrees'', ``AB = 3x - 1'').
Do not use markdown.
\par
\medskip
Respond with a single JSON object and nothing else:
\par
\{\char34{}figure\_description\char34{}: \char34{}\textless the textual
rendering\textgreater\char34{}, \char34{}confidence\char34{}: \char34{}high\char34{}
\textbar{} \char34{}medium\char34{} \textbar{} \char34{}low\char34{},
\char34{}sufficient\char34{}: true \textbar{} false\}
\par
where ``sufficient'' is true only if the correct answer is derivable from your figure description
plus the question alone.
\par
\medskip
\hrule
\medskip
USER TURN (USER\_TEMPLATE), passed as the positional prompt argument
\par
\medskip
Question that will be asked about this figure (kept verbatim in the final prompt; it may itself
contain givens you must account for):
\par
\{question\}
\par
\medskip
Correct final answer (for your completeness check only; do not reveal the answer, the solution, or
any derived value in your description):
\par
\{answer\}
\par
\medskip
Solve the problem yourself to confirm your figure description, together with the question above,
contains every fact needed to reach that answer, then produce the JSON figure description now.
\par
\end{promptbox}

\subsection{Recall calculation}

For each test diagram, we provide the Gemini 3.1 Pro judge \citep{googledeepmind2026gemini31pro} with only the reference figure description and one
model-generated candidate description. The judge first decomposes the reference figure description into independently
checkable visual facts, such as the existence of a shape, the placement of a labeled point, one
printed measure, or one marked geometric relation. Compound statements are split so that each
numeric value or relationship contributes a separate fact. It then marks each reference fact as covered
only when the candidate conveys the same information with a matching value or relationship.
Equivalent wording receives credit, whereas an omission, a contradictory value, or an incompatible
relationship does not.

Let $g_{ij}$ be the $j$th atomic fact returned for diagram $i$, and let
$c_{ij}=\mathbf{1}[g_{ij}\text{ is correctly covered}]$. The per-diagram score is
\begin{equation}
    R_i = \frac{\sum_{j=1}^{G_i}c_{ij}}{G_i},
    \qquad
    R_{\mathrm{macro}} = \frac{1}{M}\sum_{i=1}^{M}R_i,
    \label{eq:fig_desc_recall}
\end{equation}
where $G_i$ is the number of atomic reference facts and $M$ is the number of test diagrams. Thus each
diagram contributes equally regardless of how many facts its reference figure description contains.

\subsection{Gemini 3.1 Pro judge prompt}

The prompt below is used for every candidate, with \texttt{\{reference\}} and
\texttt{\{candidate\}} replaced by the corresponding descriptions.

\begin{promptbox}
\footnotesize\ttfamily\raggedright
SYSTEM INSTRUCTION\par
\medskip
You evaluate the FACTUAL RECALL of a CANDIDATE description of a geometry diagram against a REFERENCE
FIGURE DESCRIPTION of the same diagram. Treat every fact in the REFERENCE FIGURE DESCRIPTION
as correct ground truth.\par
\medskip
Step 1---Decompose the REFERENCE FIGURE DESCRIPTION into a list of atomic, independently-checkable visual
facts. Each fact is ONE self-contained piece of information, for example:\par
-- the existence of one shape (a circle, a triangle ABC, ...),\par
-- one labeled point and where it sits (P is the intersection of the two chords; A, B, C are
collinear),\par
-- ONE numeric measure or algebraic label printed on one element (segment AB = 6; angle ABC = 40
degrees; an arc labeled ``3x-1''),\par
-- ONE geometric relationship marked in the figure (AB is parallel to CD; the circle is tangent to
line L at T; M is the midpoint of AB; a right angle at C).\par
Split compound statements into separate facts. Do NOT invent facts the reference does not state. Do NOT
include the question, the task, or any solving/answer.\par
\medskip
Step 2---For EACH reference fact, decide whether the CANDIDATE description conveys the SAME information
CORRECTLY:\par
-- covered = true: the candidate asserts this fact with a matching value/relationship (wording,
order, and phrasing may differ; equivalent restatements count).\par
-- covered = false: the candidate OMITS the fact, OR states it with a DIFFERENT or CONTRADICTORY
value (e.g., the reference says a segment is 6 but the candidate says it is 8---a different value never counts
as covered).\par
\medskip
Step 3---Count how many atomic visual facts the CANDIDATE contains (every distinct fact it states,
including any that are absent from or contradict the reference).\par
\medskip
Output a single JSON object and nothing else:\par
\{\char34{}reference\_facts\char34{}: [\{\char34{}fact\char34{}:
\char34{}\textless short statement of the reference fact\textgreater\char34{},
\char34{}covered\char34{}: true \textbar{} false,
\char34{}candidate\_evidence\char34{}: \char34{}\textless short quote from the candidate, or empty
if omitted\textgreater\char34{}\}],\par
\char34{}candidate\_fact\_count\char34{}: \textless integer\textgreater,\par
\char34{}notes\char34{}: \char34{}\textless one short sentence\textgreater\char34{}\}\par
List every reference fact exactly once. candidate\_fact\_count must be an integer.\par
\medskip
\hrule
\medskip
USER TURN\par
\medskip
REFERENCE FIGURE DESCRIPTION:\par
\{reference\}\par
\medskip
CANDIDATE description (to be scored against the reference):\par
\{candidate\}\par
\medskip
Decompose the reference into atomic facts, mark which ones the candidate conveys correctly, and count
the candidate's total facts. Respond with the JSON object now.\par
\end{promptbox}

\section{Additional Experimental Results}
\label{sec:detailed_perception_results}

\subsection{Figure-description recall standard deviations}

The main tables report mean figure-description recall over three candidate descriptions sampled
independently for every test diagram at temperature 0.8 using seeds 1001, 1002, and 1003. Each
candidate is scored against the same reference figure description using the judge in
Section~\ref{sec:fig_desc_recall}. Because space permits only the mean in the main tables,
Table~\ref{tab:figdesc_repeated_qwen3} and Table~\ref{tab:figdesc_repeated_qwen25} repeat those means
with their standard deviations across the three sampled-description draws.

\begin{table*}[t]
    \centering
    \caption{Figure-description recall for Qwen3-VL-2B and Qwen3-VL-8B students. Entries are mean
    $\pm$ standard deviation across three independently sampled descriptions per diagram; the main
    tables report the corresponding means. Per column, the best result is $\bm{bold}$ and the second
    best is \underline{underlined}.}
    \label{tab:figdesc_repeated_qwen3}
    \small
    \begin{tabular}{@{}lcc@{}}
        \toprule
        Method & 2B student & 8B student \\
        \midrule
        Base                                      & $0.567\pm0.007$ & $0.687\pm0.002$ \\
        SFT (rejection sampling)                  & $0.605\pm0.009$ & $0.781\pm0.008$ \\
        SFT (describe-then-solve)                 & $0.670\pm0.010$ & $0.798\pm0.011$ \\
        SFT (describe-figure)                     & $\underline{0.770\pm0.005}$ & $0.839\pm0.007$ \\
        SFT (mix: rejection + describe)           & $\bm{0.784\pm0.006}$ & $\underline{0.846\pm0.003}$ \\
        GRPO (plain prompt)                       & $0.552\pm0.009$ & $0.676\pm0.007$ \\
        GRPO (figure-desc prompt)                 & $0.415\pm0.005$ & $0.668\pm0.011$ \\
        GRPO (mix: plain + figure-desc)           & $0.560\pm0.003$ & $0.708\pm0.005$ \\
        Naive OPD (larger teacher)                & $0.655\pm0.005$ & $0.750\pm0.013$ \\
        Privileged OPD (larger teacher)           & $0.742\pm0.004$ & $\bm{0.878\pm0.003}$ \\
        Minimal-edit (larger teacher, ours)       & $0.633\pm0.003$ & $0.727\pm0.005$ \\
        Minimal-edit (self-teacher, ours)         & $0.588\pm0.004$ & $0.745\pm0.003$ \\
        \bottomrule
    \end{tabular}
\end{table*}

\begin{table}[t]
    \centering
    \caption{Figure-description recall on the Qwen2.5-VL-7B backbone. Entries are mean $\pm$
    standard deviation across three independently sampled descriptions per diagram; the main table
    reports the corresponding means. The best result is $\bm{bold}$ and the second best is
    \underline{underlined}.}
    \label{tab:figdesc_repeated_qwen25}
    \small
    \begin{tabular}{@{}lc@{}}
        \toprule
        Model & Recall \\
        \midrule
        Qwen2.5-VL-7B                   & $0.431\pm0.007$ \\
        Perception-R1-7B                & $\bm{0.541\pm0.002}$ \\
        NoisyRollout-7B                 & $\underline{0.500\pm0.014}$ \\
        PAPO-7B                         & $0.470\pm0.005$ \\
        DeepEyes-7B                     & $0.447\pm0.003$ \\
        Vision-R1-7B                    & $0.486\pm0.007$ \\
        Minimal-edit (ours)             & $0.477\pm0.009$ \\
        \bottomrule
    \end{tabular}
\end{table}

\subsection{Geoperception breakdown by predicate}

Tables~\ref{tab:geoperception_predicate_2b}--\ref{tab:geoperception_predicate_7b} decompose
the aggregate Geoperception pass@1 reported in the main tables by predicate category, with eight
rollouts per example. All three tables report means only so that every row and model scale uses a
consistent format. The predicate columns are angle classification
(Angle), line comparison (Line), point on line (OnLine), equality (Equal), point on circle
(OnCirc), parallelism (Parallel), and perpendicularity (Perp).

\begin{table*}[t]
    \centering
    \caption{Geoperception pass@1 by predicate for Qwen3-VL-2B students. Per column, the best
    result is $\bm{bold}$ and the second best is \underline{underlined}.}
    \label{tab:geoperception_predicate_2b}
    \scriptsize
    \setlength{\tabcolsep}{2.5pt}
    \resizebox{\textwidth}{!}{%
    \begin{tabular}{@{}lcccccccc@{}}
        \toprule
        Method & Angle & Line & OnLine & Equal & OnCirc & Parallel & Perp & Avg. \\
        \midrule
        Base & $0.129$ & $0.158$ & $0.093$ & $0.191$ & $0.405$ & $0.128$ & $0.147$ & $0.179$ \\
        SFT (rejection sampling) & $0.165$ & $0.231$ & $0.088$ & $0.254$ & $0.421$ & $0.360$ & $0.188$ & $0.244$ \\
        SFT (describe-then-solve) & $0.346$ & $0.423$ & $0.241$ & $0.277$ & $0.450$ & $0.511$ & $0.284$ & $0.362$ \\
        SFT (describe-figure) & $0.341$ & $0.307$ & $0.207$ & $\bm{0.442}$ & $0.355$ & $\underline{0.608}$ & $0.253$ & $0.359$ \\
        SFT (mix) & $0.172$ & $0.248$ & $0.168$ & $0.276$ & $0.413$ & $0.527$ & $0.217$ & $0.289$ \\
        GRPO (plain prompt) & $0.262$ & $0.290$ & $0.217$ & $0.243$ & $0.392$ & $0.287$ & $0.262$ & $0.279$ \\
        GRPO (figure-desc prompt) & $0.447$ & $0.461$ & $0.149$ & $0.189$ & $0.259$ & $0.233$ & $0.208$ & $0.278$ \\
        GRPO (mixed prompts) & $0.475$ & $0.478$ & $0.206$ & $0.264$ & $0.107$ & $0.383$ & $0.246$ & $0.308$ \\
        Naive OPD (8B teacher) & $0.207$ & $0.283$ & $0.204$ & $0.249$ & $\underline{0.575}$ & $0.393$ & $0.259$ & $0.310$ \\
        Privileged OPD (8B teacher) & $0.195$ & $0.296$ & $0.172$ & $0.281$ & $0.435$ & $\bm{0.650}$ & $0.209$ & $0.320$ \\
        Minimal-edit (8B teacher) & $\underline{0.511}$ & $\bm{0.605}$ & $\underline{0.248}$ & $0.311$ & $\bm{0.580}$ & $0.557$ & $\bm{0.372}$ & $\bm{0.455}$ \\
        Minimal-edit (self-teacher) & $\bm{0.535}$ & $\underline{0.601}$ & $\bm{0.252}$ & $\underline{0.329}$ & $0.504$ & $0.507$ & $\underline{0.340}$ & $\underline{0.438}$ \\
        \bottomrule
    \end{tabular}
    }
\end{table*}

\begin{table*}[t]
    \centering
    \caption{Geoperception pass@1 by predicate for Qwen3-VL-8B students. Per column, the best
    result is $\bm{bold}$ and the second best is \underline{underlined}.}
    \label{tab:geoperception_predicate_8b}
    \scriptsize
    \setlength{\tabcolsep}{2.5pt}
    \resizebox{\textwidth}{!}{%
    \begin{tabular}{@{}lcccccccc@{}}
        \toprule
        Method & Angle & Line & OnLine & Equal & OnCirc & Parallel & Perp & Avg. \\
        \midrule
        Base & $0.349$ & $0.454$ & $0.004$ & $0.477$ & $0.755$ & $0.615$ & $0.406$ & $0.437$ \\
        SFT (rejection sampling) & $0.345$ & $0.408$ & $0.033$ & $0.448$ & $0.693$ & $0.645$ & $0.398$ & $0.424$ \\
        SFT (describe-then-solve) & $0.468$ & $0.595$ & $\underline{0.128}$ & $0.459$ & $0.702$ & $0.720$ & $0.428$ & $0.500$ \\
        SFT (describe-figure) & $0.356$ & $0.418$ & $\bm{0.137}$ & $0.547$ & $0.750$ & $0.712$ & $0.309$ & $0.461$ \\
        SFT (mix) & $0.376$ & $0.466$ & $0.110$ & $0.467$ & $0.767$ & $0.761$ & $0.445$ & $0.484$ \\
        GRPO (plain prompt) & $\bm{0.634}$ & $\bm{0.754}$ & $0.012$ & $0.546$ & $0.733$ & $0.590$ & $0.443$ & $0.530$ \\
        GRPO (figure-desc prompt) & $0.537$ & $0.598$ & $0.028$ & $\bm{0.605}$ & $0.739$ & $0.625$ & $0.406$ & $0.506$ \\
        GRPO (mixed prompts) & $\underline{0.584}$ & $0.616$ & $0.000$ & $0.469$ & $0.760$ & $0.696$ & $0.377$ & $0.500$ \\
        Naive OPD (32B teacher) & $0.418$ & $0.439$ & $0.007$ & $0.388$ & $\underline{0.772}$ & $\underline{0.798}$ & $0.438$ & $0.466$ \\
        Privileged OPD (32B teacher) & $0.399$ & $0.467$ & $0.013$ & $0.466$ & $\bm{0.791}$ & $\bm{0.849}$ & $\bm{0.489}$ & $0.496$ \\
        Minimal-edit (32B teacher) & $0.567$ & $\underline{0.717}$ & $0.016$ & $\underline{0.588}$ & $0.762$ & $0.765$ & $\underline{0.459}$ & $\bm{0.553}$ \\
        Minimal-edit (self-teacher) & $0.582$ & $0.715$ & $0.037$ & $0.562$ & $0.759$ & $0.743$ & $0.453$ & $\underline{0.550}$ \\
        \bottomrule
    \end{tabular}
    }
\end{table*}

\begin{table*}[t]
    \centering
    \caption{Geoperception pass@1 by predicate on the Qwen2.5-VL-7B backbone. Per column, the best
    result is $\bm{bold}$ and the second best is \underline{underlined}.}
    \label{tab:geoperception_predicate_7b}
    \scriptsize
    \setlength{\tabcolsep}{2.5pt}
    \resizebox{\textwidth}{!}{%
    \begin{tabular}{@{}lcccccccc@{}}
        \toprule
        Model & Angle & Line & OnLine & Equal & OnCirc & Parallel & Perp & Avg. \\
        \midrule
        Qwen2.5-VL-7B & $0.382$ & $0.475$ & $0.178$ & $0.258$ & $0.404$ & $0.365$ & $0.247$ & $0.330$ \\
        Perception-R1-7B & $\underline{0.526}$ & $\bm{0.600}$ & $0.170$ & $\underline{0.293}$ & $\bm{0.568}$ & $\bm{0.536}$ & $0.248$ & $\bm{0.420}$ \\
        NoisyRollout-7B & $0.490$ & $\underline{0.589}$ & $\underline{0.218}$ & $0.273$ & $\underline{0.507}$ & $0.506$ & $\underline{0.287}$ & $\underline{0.410}$ \\
        PAPO-7B & $0.453$ & $0.571$ & $0.162$ & $0.249$ & $0.465$ & $0.334$ & $0.223$ & $0.351$ \\
        DeepEyes-7B & $0.442$ & $0.569$ & $0.155$ & $0.233$ & $0.478$ & $0.402$ & $0.265$ & $0.363$ \\
        Vision-R1-7B & $\bm{0.534}$ & $0.573$ & $0.132$ & $\bm{0.309}$ & $0.488$ & $0.426$ & $0.279$ & $0.392$ \\
        Minimal-edit (ours) & $0.480$ & $0.544$ & $\bm{0.263}$ & $0.278$ & $0.473$ & $\underline{0.531}$ & $\bm{0.291}$ & $0.409$ \\
        \bottomrule
    \end{tabular}
    }
\end{table*}

\section{Teacher Prompts}
\label{sec:teacher_prompts}

\subsection{Privileged teacher prompt}
\label{sec:privileged_teacher_prompt}

The conventional privileged teacher is conditioned on the diagram image, the collected figure
description, and the original question. The student's rollout is then appended as the
teacher-forced sequence whose token probabilities are scored.

\begin{promptbox}
\footnotesize\ttfamily\raggedright
\textless image\textgreater
\par
\{figure description\}
\par
\{question\}
\par
\end{promptbox}

\subsection{Minimal-edit teacher prompt}
\label{sec:minimal_edit_prompt}

For each incorrect student rollout, the frozen teacher is conditioned on the following prompt.
During training, the teacher does not generate the requested rewrite; the prompt conditions a
teacher-forced scoring pass over the student's existing tokens. The rewrite instruction defines the
minimal-edit distribution against which the token-level $k_1$ estimates are computed.

\begin{promptbox}
\footnotesize\ttfamily\raggedright
\textless image\textgreater\par
\{figure description\}\par
\{question\}\par
\medskip
--- REFERENCE CORRECT SOLUTION ---\par
\{ref\}\par
--- END REFERENCE SOLUTION ---\par
\medskip
--- STUDENT SOLUTION ---\par
\{student\}\par
--- END STUDENT SOLUTION ---\par
\medskip
Rewrite the student solution above into a correct solution by making the SMALLEST possible change:
copy it VERBATIM and alter only the specific spans that are actually wrong, keeping every correct
sentence, number, and symbol exactly as written, in the student's original style and voice. Write
ONLY the corrected solution itself as a single self-contained solution that reads as if it were
correct from the start. Do NOT refer to ``the student'', do not state that anything was wrong,
missing, changed, or corrected, and add no commentary. End with the final answer in
\textbackslash{}boxed\{\}.
\end{promptbox}

\section{Qualitative Examples}
\label{subsec:additional_k1}

Figure~\ref{fig:k1_visual} illustrates how minimal-edit conditioning concentrates corrective-token
selection around a perceptual error. We provide two additional examples spanning a downstream
arithmetic mistake and a second diagram-reading
error. In each visualization, both teachers score the same fixed student rollout; only their
conditioning differs. Red tokens have $k_t>0$ and would enter the corrective mask, whereas green
tokens have $k_t\leq0$ and are discarded.

To make the behavior induced by the minimal-edit prompt more concrete, we also show a hypothetical
minimal-edit rollout for each example. These edited responses are generated only for exposition.
During training, the minimal-edit teacher is never rolled out: it receives the description,
reference solution, and fixed student response, then scores the tokens of that existing response in
a teacher-forced pass. The textual edits shown below are therefore neither training targets nor
inputs to the loss. They illustrate what the teacher could generate under the same minimal-edit
prompt, whereas the actual corrective mask is determined solely by the token-level $k_1$ estimates
on the unchanged student rollout.

\begin{figure*}
    \centering
    \includegraphics[width=\linewidth]{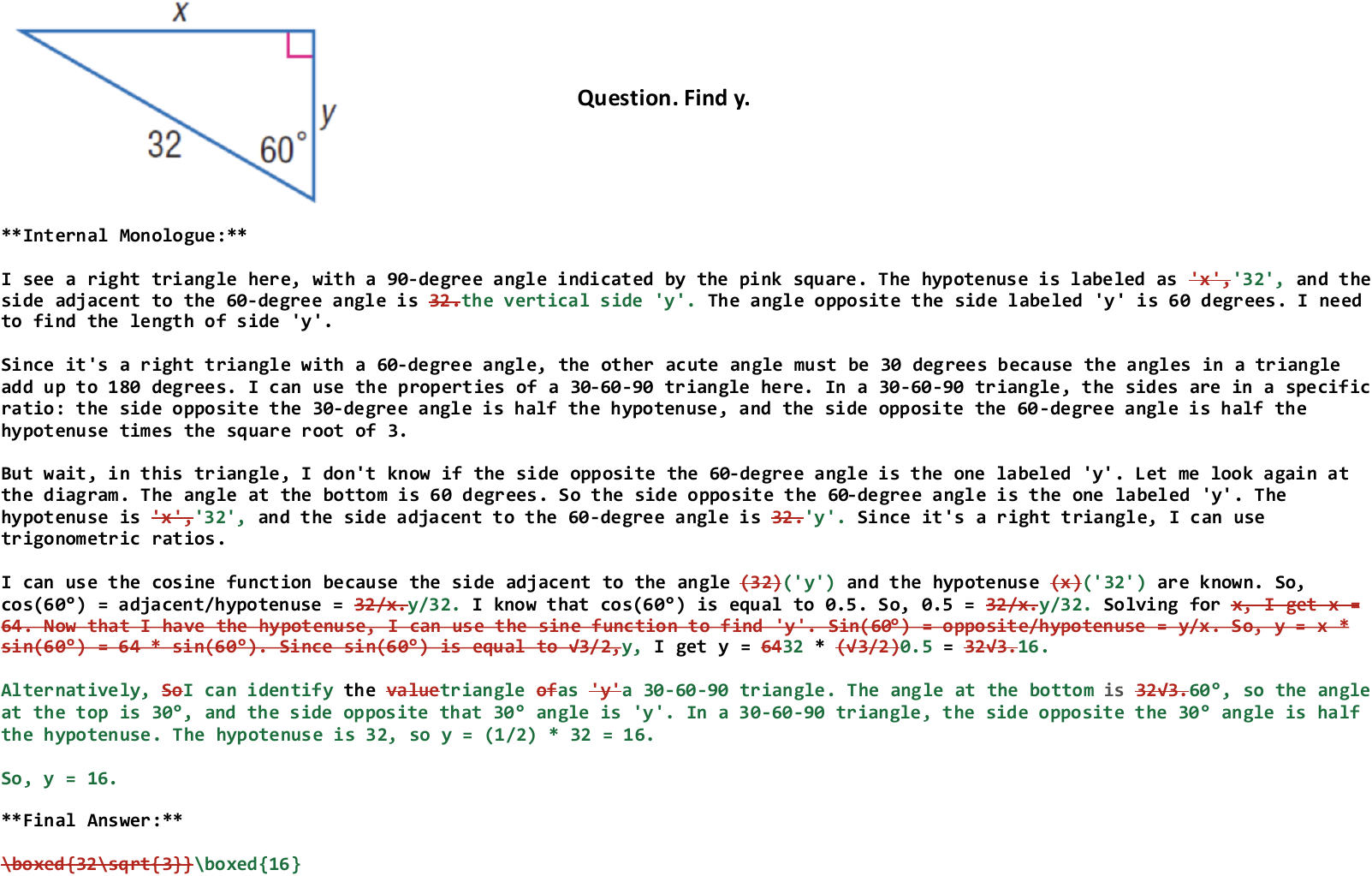}
    \caption{Hypothetical minimal-edit rollout for the leg--hypotenuse example in
    Figure~\ref{fig:k1_visual}. Red strikethrough marks text removed from the student response, and
    green marks inserted text. The edit identifies 32 as the hypotenuse, revises the dependent
    reasoning, and changes the answer from $32\sqrt{3}$ to 16 while preserving text that remains
    applicable. This rollout is generated only for visualization; training does not decode an
    edited response and instead scores the original student tokens under the minimal-edit prompt.}
    \label{fig:minimal_edit_rollout1}
\end{figure*}

Figure~\ref{fig:minimal_edit_rollout1} shows the hypothetical edit corresponding to
Figure~\ref{fig:k1_visual}. The edit corrects the initial assignment of the side labeled 32 and the
hypotenuse, removes the derivation that follows from that misreading, and changes the final answer
from $32\sqrt{3}$ to 16. The retained text demonstrates the prompt's preservation objective, while
the revision makes explicit the correction suggested by the teacher-disfavored positions.

\begin{figure*}
    \centering
    \includegraphics[width=\linewidth]{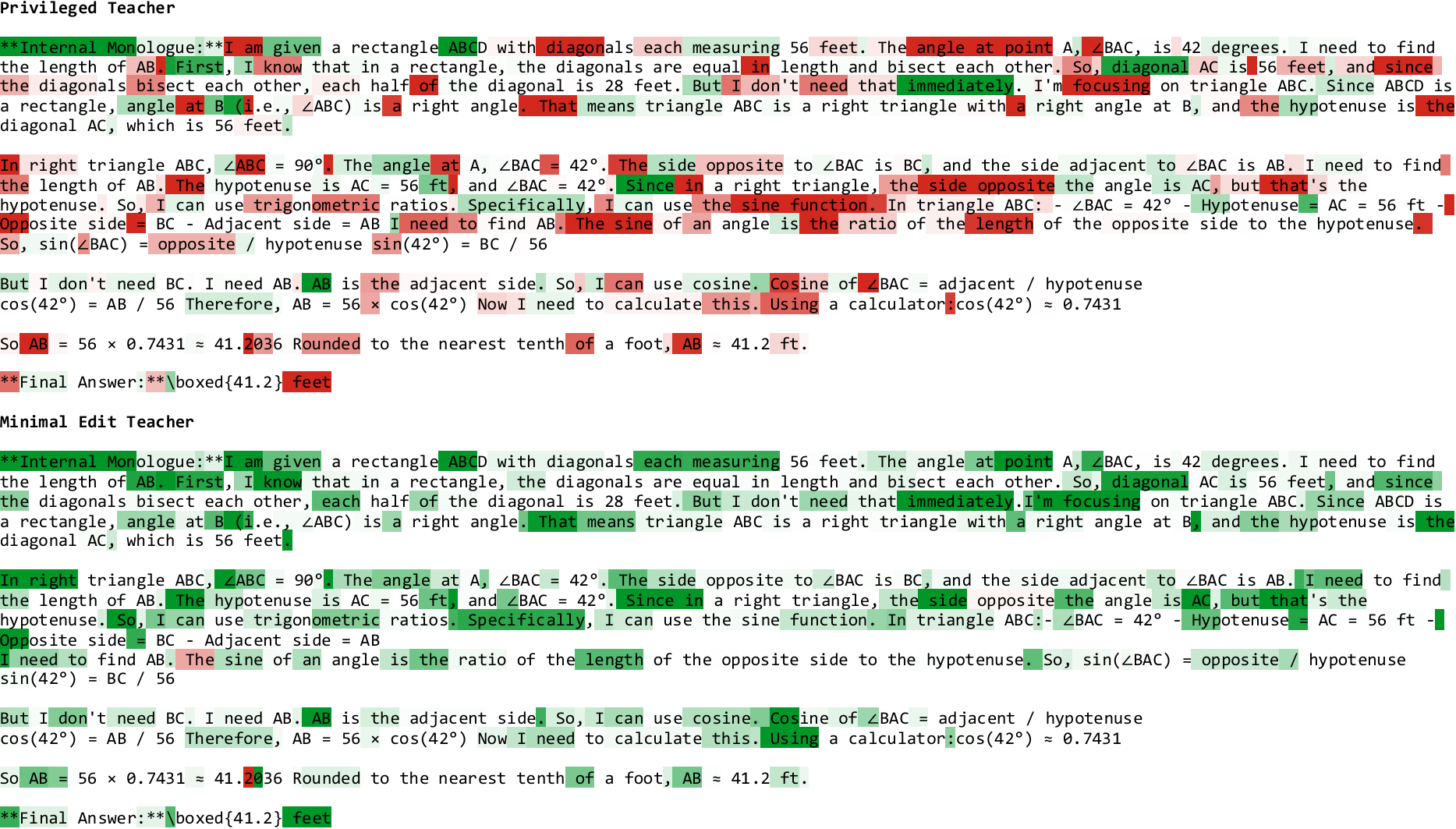}
    \caption{Edit conditioning concentrates corrective-token selection near a numerical error. Token-level $k_1$ estimates for an incorrect
    solution that selects the correct relation, $AB=56\cos(42^\circ)$, but evaluates it incorrectly
    as 41.2 rather than 41.6. The conventional privileged teacher assigns a positive $k_1$ estimate to
    many otherwise acceptable wording choices (top), whereas minimal-edit conditioning assigns high
    probability to most of the student's derivation and concentrates the corrective signal near the erroneous
    calculation (bottom). Colors follow Figure~\ref{fig:k1_visual}.}
    \label{fig:k1_visual2}
\end{figure*}

Figure~\ref{fig:k1_visual2} considers a response whose geometric setup is correct but whose final
arithmetic is not. The student identifies $AB$ as the side adjacent to the $42^\circ$ angle and
correctly writes $AB=56\cos(42^\circ)$, but evaluates the product as $41.2$ rather than $41.6$.
The conventional privileged teacher produces positive $k_1$ estimates throughout the otherwise
usable derivation. By contrast, the minimal-edit teacher assigns at least as much probability as
the student to most of those tokens and concentrates its positive estimate near the incorrect
multiplication. Thus, this corrective-token selection pattern is not limited to perceptual
mistakes: it also concentrates selected positions near a downstream numerical error while
preserving the correct setup that precedes it.

\begin{figure*}
    \centering
    \includegraphics[width=\linewidth]{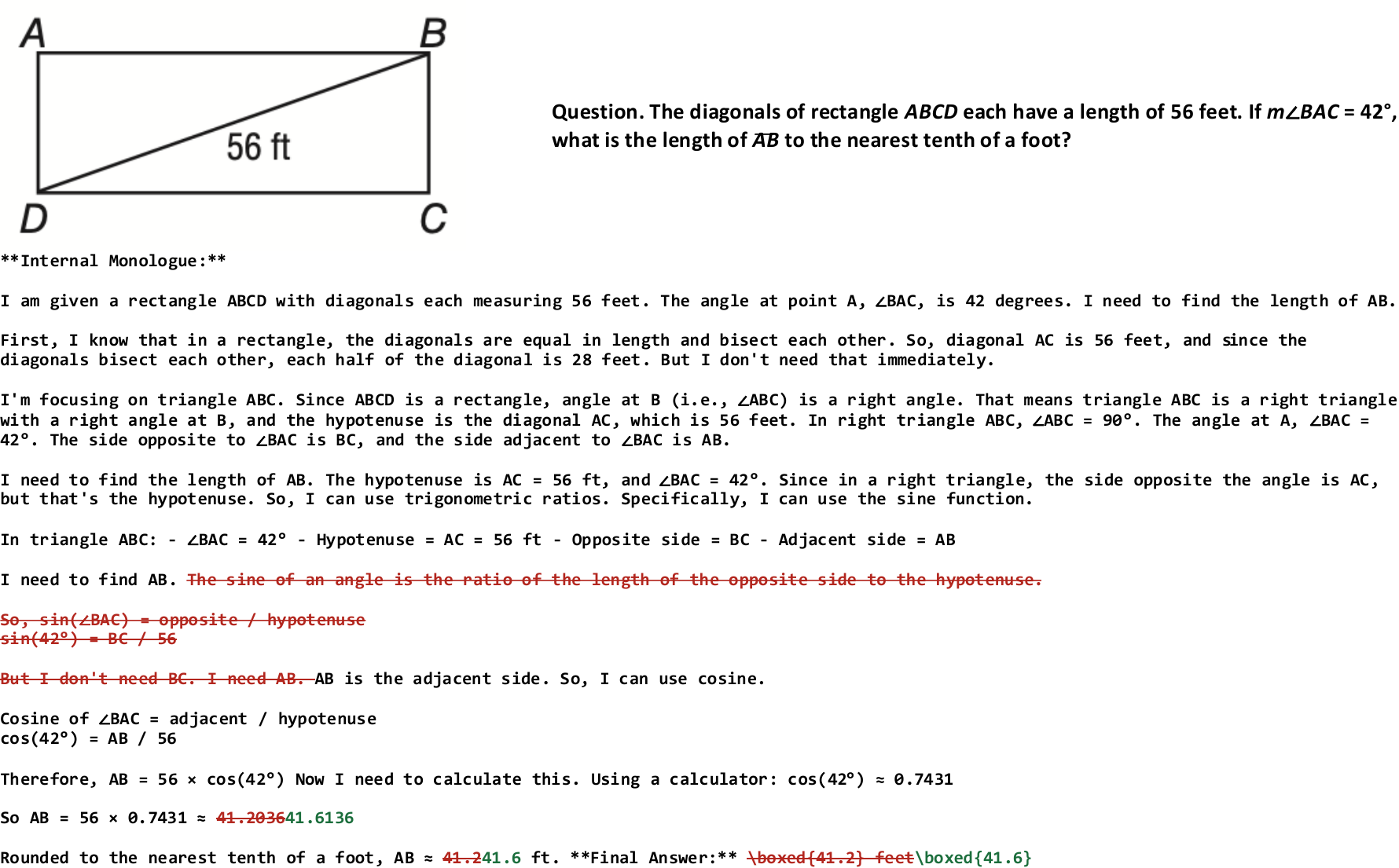}
    \caption{Hypothetical minimal-edit rollout for the arithmetic-error example in
    Figure~\ref{fig:k1_visual2}. Red strikethrough marks removed text, and green marks inserted text.
    The edit preserves $AB=56\cos(42^\circ)$, corrects the product from 41.2036 to 41.6136, and
    changes the rounded answer from 41.2 to 41.6 feet. This rollout is generated only for
    visualization; the training procedure scores the fixed student response without decoding this
    revision.}
    \label{fig:minimal_edit_rollout2}
\end{figure*}

The hypothetical edit in Figure~\ref{fig:minimal_edit_rollout2} preserves the correct geometric
setup and cosine relation, removes an unnecessary detour through sine, and corrects the product and
rounded answer. This response-level view agrees with the token-level visualization: the substantive
correction is numerical rather than geometric.

\begin{figure*}
    \centering
    \includegraphics[width=\linewidth]{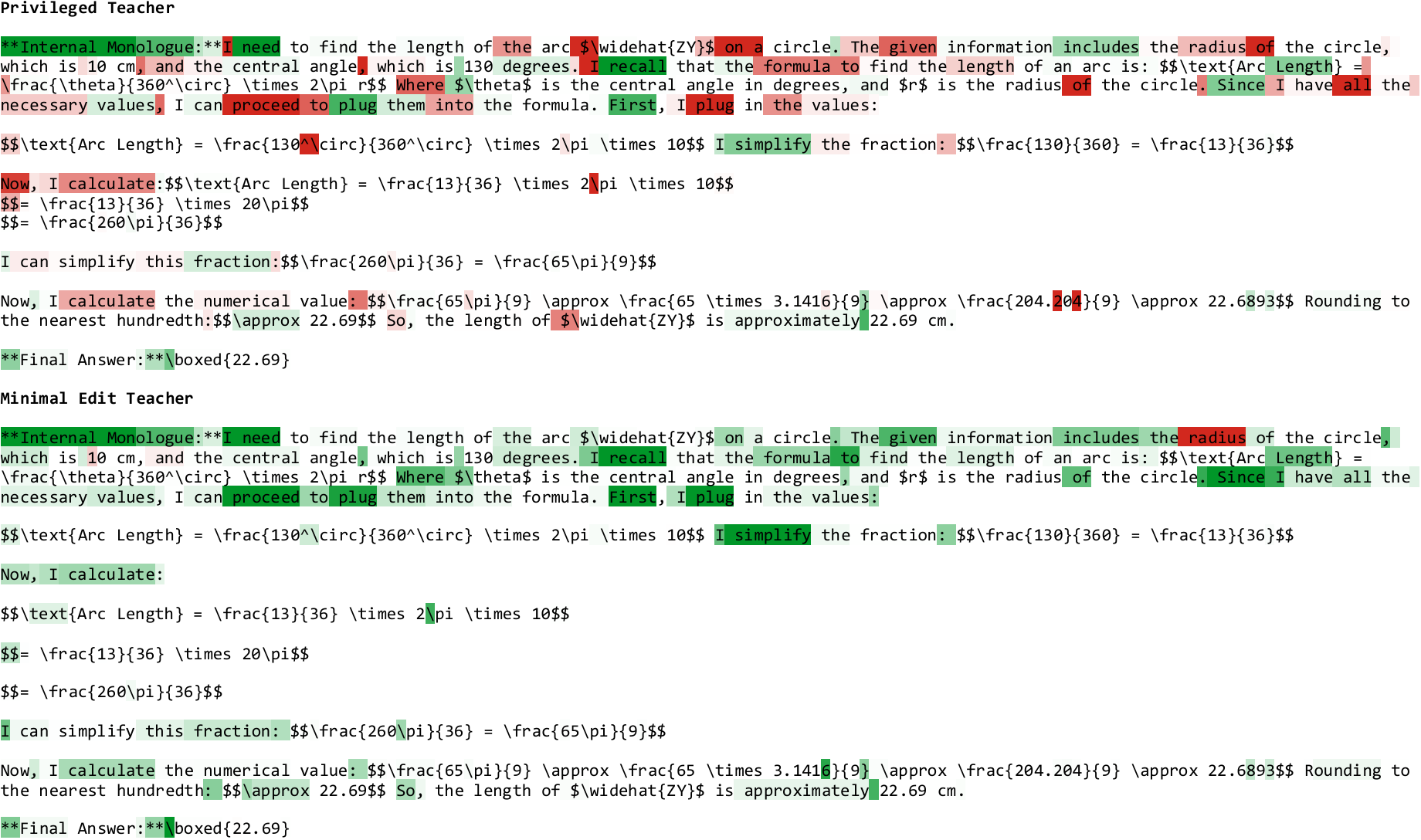}
    \caption{Edit conditioning concentrates corrective-token selection near a diameter--radius error. The student misreads the labeled
    diameter of 10 cm as the radius and consequently computes the arc length using $r=10$. The
    conventional privileged teacher assigns positive $k_1$ estimates throughout the response
    (top), whereas the minimal-edit teacher concentrates the corrective signal on the tokens that
    identify 10 cm as the radius (bottom). Colors follow Figure~\ref{fig:k1_visual}.}
    \label{fig:k1_visual3}
\end{figure*}

Figure~\ref{fig:k1_visual3} presents a second perceptual failure. The diagram labels a diameter of
10 cm, but the student treats that value as the radius. Its subsequent arc-length derivation is
consistent with this incorrect premise, producing $65\pi/9\approx22.69$ instead of the value obtained
with $r=5$. A conventional privileged teacher assigns positive $k_1$ estimates across much of the
response, including the downstream calculation. The minimal-edit teacher instead preserves most of
that calculation and focuses its strongest corrective signal on the initial identification of the
radius. This is the desired behavior: correcting the misread quantity makes the otherwise coherent
derivation reusable rather than treating every consequence of the initial error as an independent
mistake.

\begin{figure*}
    \centering
    \includegraphics[width=\linewidth]{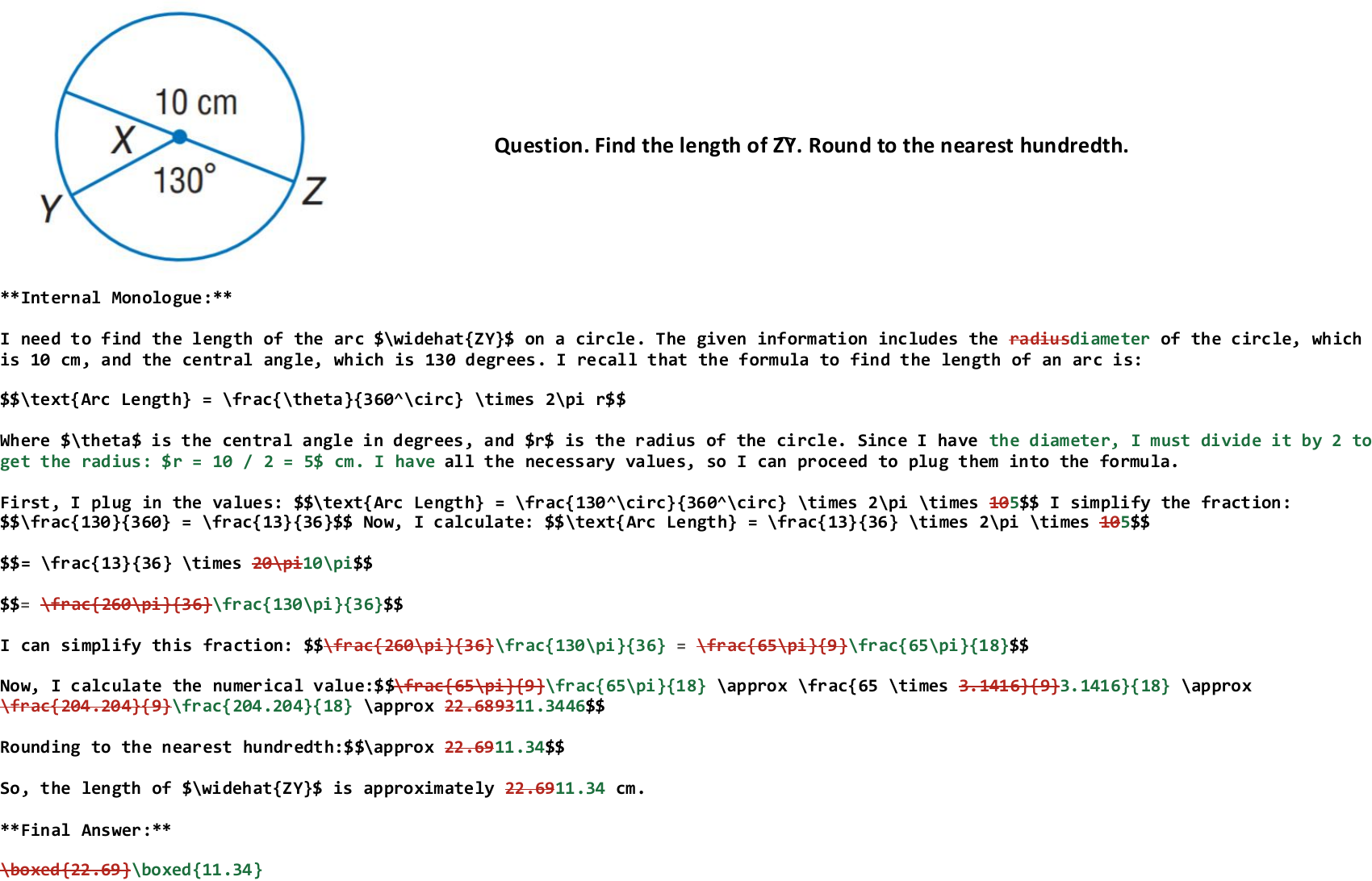}
    \caption{Hypothetical minimal-edit rollout for the diameter--radius example in
    Figure~\ref{fig:k1_visual3}. Red strikethrough marks removed text, and green marks inserted text.
    The edit recognizes 10 cm as the diameter, sets $r=5$, and updates the arc length from
    $65\pi/9\approx22.69$ to $65\pi/18\approx11.34$. This rollout is generated only for
    visualization; during training, the teacher generates no revised response and only scores the
    original student tokens.}
    \label{fig:minimal_edit_rollout3}
\end{figure*}

Figure~\ref{fig:minimal_edit_rollout3} shows the corresponding hypothetical edit. It replaces
``radius'' with ``diameter,'' introduces the conversion $r=10/2=5$, and updates the subsequent
calculation and final answer to $65\pi/18\approx11.34$. The edit makes visible how a localized
perceptual correction propagates through an otherwise valid solution structure.

Together with Figure~\ref{fig:k1_visual}, these examples cover three regimes: a localized
misidentification of a triangle's leg and hypotenuse, a localized arithmetic error after a correct
setup, and a diameter--radius reading error whose consequences propagate through an otherwise valid
derivation. Across all three,
conventional privileged conditioning responds to broad distributional differences, whereas
minimal-edit conditioning better preserves tokens that do not need to change. The corrective mask
then converts that response-specific preference into a sparse training signal.

\end{document}